%% file: sn-article.tex
\documentclass[sn-mathphys]{sn-jnl}

\jyear{2022}%

\theoremstyle{thmstyleone}%
\theoremstyle{thmstyletwo}%

\theoremstyle{thmstylethree}%

\usepackage[utf8]{inputenc} 
\usepackage{amsfonts}       
\usepackage{nicefrac}       
\usepackage{microtype}      
\usepackage{xcolor}         
\usepackage{blindtext}
\usepackage{graphicx}
\usepackage{subcaption}
\usepackage{amsmath}
\usepackage{multicol}
\usepackage{algorithm}
\usepackage{algpseudocode}
\usepackage{listings}
\usepackage[export]{adjustbox}
\usepackage{graphics}
\usepackage{array}

\definecolor{codegreen}{rgb}{0,0.6,0}
\definecolor{codegray}{rgb}{0.5,0.5,0.5}
\definecolor{codepurple}{rgb}{0.58,0,0.82}
\definecolor{backcolour}{rgb}{0.95,0.95,0.92}

\lstdefinestyle{mystyle}{
    backgroundcolor=\color{backcolour},   
    commentstyle=\color{codegreen},
    keywordstyle=\color{magenta},
    numberstyle=\tiny\color{codegray},
    stringstyle=\color{codepurple},
    basicstyle=\ttfamily\footnotesize,
    breakatwhitespace=false,         
    breaklines=true,                 
    captionpos=b,                    
    keepspaces=true,                 
    numbers=left,                    
    numbersep=5pt,                  
    showspaces=false,                
    showstringspaces=false,
    showtabs=false,                  
    tabsize=2
}

\begin{document}

\title[Automatic MILP Solver Configuration By Learning Problem Similarities]{Automatic MILP Solver Configuration By Learning Problem Similarities}


\author*[1]{\fnm{Abdelrahman} \sur{Hosny}}\email{abdelrahman\_hosny@brown.edu}

\author[1,2]{\fnm{Sherief} \sur{Reda}}\email{sherief\_reda@brown.edu}

\affil[1]{\orgdiv{Department of Computer Science}, \orgname{Brown University}, \orgaddress{ \city{Providence}, \postcode{02912}, \state{RI}, \country{USA}}}

\affil[2]{\orgdiv{School of Engineering}, \orgname{Brown University}, \orgaddress{\city{Providence}, \postcode{02912}, \state{RI}, \country{USA}}}


\abstract{
A large number of real-world optimization problems can be formulated as Mixed Integer Linear Programs (MILP).
MILP solvers expose numerous configuration parameters to control their internal algorithms.
Solutions, and their associated costs or runtimes, are significantly affected by the choice of the configuration parameters, even when problem instances have the same number of decision variables and constraints.
On one hand, using the default solver configuration leads to suboptimal solutions.
On the other hand, searching and evaluating a large number of configurations for every problem instance is time-consuming and, in some cases, infeasible.
In this study, we aim to predict configuration parameters for unseen problem instances that yield lower-cost solutions without the time overhead of searching-and-evaluating configurations at the solving time.
Toward that goal, we first investigate the cost correlation of MILP problem instances that come from the same distribution when solved using different configurations. We show that instances that have similar costs using one solver configuration also have similar costs using another solver configuration in the same runtime environment.
After that, we present a methodology based on Deep Metric Learning to learn MILP similarities that correlate with their final solutions' costs.
At inference time, given a new problem instance, it is first projected into the learned metric space using the trained model, and configuration parameters are instantly predicted using previously-explored configurations from the nearest neighbor instance in the learned embedding space.
Empirical results on real-world problem benchmarks show that our method predicts configuration parameters that improve solutions' costs by up to 38\% compared to existing approaches.
}


\pacs[Conflict of Interest]{Abdelrahman Hosny declares that he has no conflict
of interest. Sherief Reda declares that he has no conflict of interest.}


\maketitle

\section{Introduction}
\label{sec:introduction}
\input{1_intro}

\section{Related Work}
\label{sec:related-work}
\input{2_related_work}

\section{Preliminaries}
\label{sec:preliminaries}
\input{3_prelim}

\section{Data Validation}
\label{sec:motivation}
\input{4_motivation}

\section{Methodology}
\label{sec:methodology}
\input{4_method}

\section{Results}
\label{sec:expr}
\input{5_expr}

\section{Discussion}
\label{sec:disucssion}
\input{6_discussion}

\section{Conclusions and Future Work}
\label{sec:conclusion}
\input{7_conclusion}

\bmhead{Supplementary information}

Appendix \ref{appendix:data-management} presents details on the dataset and configuration parameters used in this study. The source code is available at this link\footnote{Link: \url{https://drive.google.com/file/d/15YGwH2o1CVBXFufe35hrOntv08YT2Ux9/view?usp=share_link}}, and the trained model and the data store are available at this link\footnote{URL: \url{https://drive.google.com/file/d/1-qzBym0TBsfk4WuemB9ffuTyvFNY5s7u/view?usp=share_link}}.

\section*{Compliance with Ethical Standards}

\bmhead{Funding}
This study was partially funded by NSF grant 1814920 and DoD ARO grant W911NF-19-1-0484.

\bmhead{Conflict of Interest} 
Abdelrahman Hosny declares that he has no conflict of interest.
Sherief Reda declares that he has no conflict of interest.

\bmhead{Ethical approval} This article does not contain any studies with human participants or animals performed by any of the authors.

\bibliography{sn-bibliography}

\newpage
\begin{appendices}

\section{Data Management}
\label{appendix:data-management}
\input{appendix_1}




\end{appendices}




\end{document}

%% file: 1_intro.tex
Mixed Integer Linear Programs (MILP) is a class of NP-hard problems where the goal is to minimize a linear objective function subject to linear constraints, with some or all decision variables restricted to integer or binary values \citep{floudas2005mixed}.
This formulation has applications in numerous fields, such as transportation, retail, manufacturing and management \citep{paschos2014applications, becker2021extending}. 
For example, last-mile delivery companies repeatedly solve the vehicle routing problem as daily delivery tasks (stops and routes) change, with the goal of minimizing total delivery costs\nobreakspace\citep{louati2021mixed}.
Similarly, crew scheduling problems have to be solved daily or weekly in the aviation industry, where the MILP formulation is the most practical notation for expressing such problems \citep{deveci2018survey}.
Over the years, solvers have been well researched and practically engineered to address these problems, such as SCIP\nobreakspace\citep{gamrath2020scip}, CPLEX\nobreakspace\citep{manual1987ibm}, and Gurobi \citep{bixby2007gurobi}.
These solvers mostly use branch-and-bound methods combined with heuristics to direct the search process for solving a MILP \citep{achterberg2007constraint}.
In order to tune their behavior, they expose a large number of configuration parameters that control the search trajectory.
In implementing a branching strategy, SCIP, as a widely-used open-source solver, exposes configuration parameters to help in selecting the most promising decision variable to branch on at each node in the branch-and-bound tree, which significantly impact the efficiency and effectiveness of the solver.
For example, the branching score function, $branching/scorefunc \in \{p, s , q\}$, and the branching score factor, $branching/scorefac \in \{0, 1\}$, help evaluate the potential of expanding a specific branch in the search tree.
Those are just two of more than 2500 parameters with integer, continuous or categorical configuration spaces.

Automatic algorithm configuration is the task of identifying optimal parameter configurations for solving unseen problem instances by training on a collection of representative problem instances \citep{eryoldacs2022literature}.
This process can be divided into two distinct phases.
The primary tuning phase involves selecting a parameters configuration based on a set of training instances representative of a specific problem.
Subsequently, during the testing phase, the chosen parameter configuration is employed to tackle unseen instances of the same problem.
The objective is to identify, during the tuning phase, a parameters configuration that minimizes a particular cost metric over the set of instances that will be encountered during the deployment phase. 
There exist a rich literature on efficient search methods for automatic algorithm configuration for optimization problems \citep{lopez2016irace, birattari2009tuning, birattari2002racing, maron1997racing, hoos2012automated, kerschke2019automated}.
They mostly differ in how they navigate the huge search space to find potential configurations fast during the tuning phase.

\begin{figure}[t]
    \centering
    \begin{subfigure}{0.5\textwidth}
        \includegraphics[clip, scale=0.25]{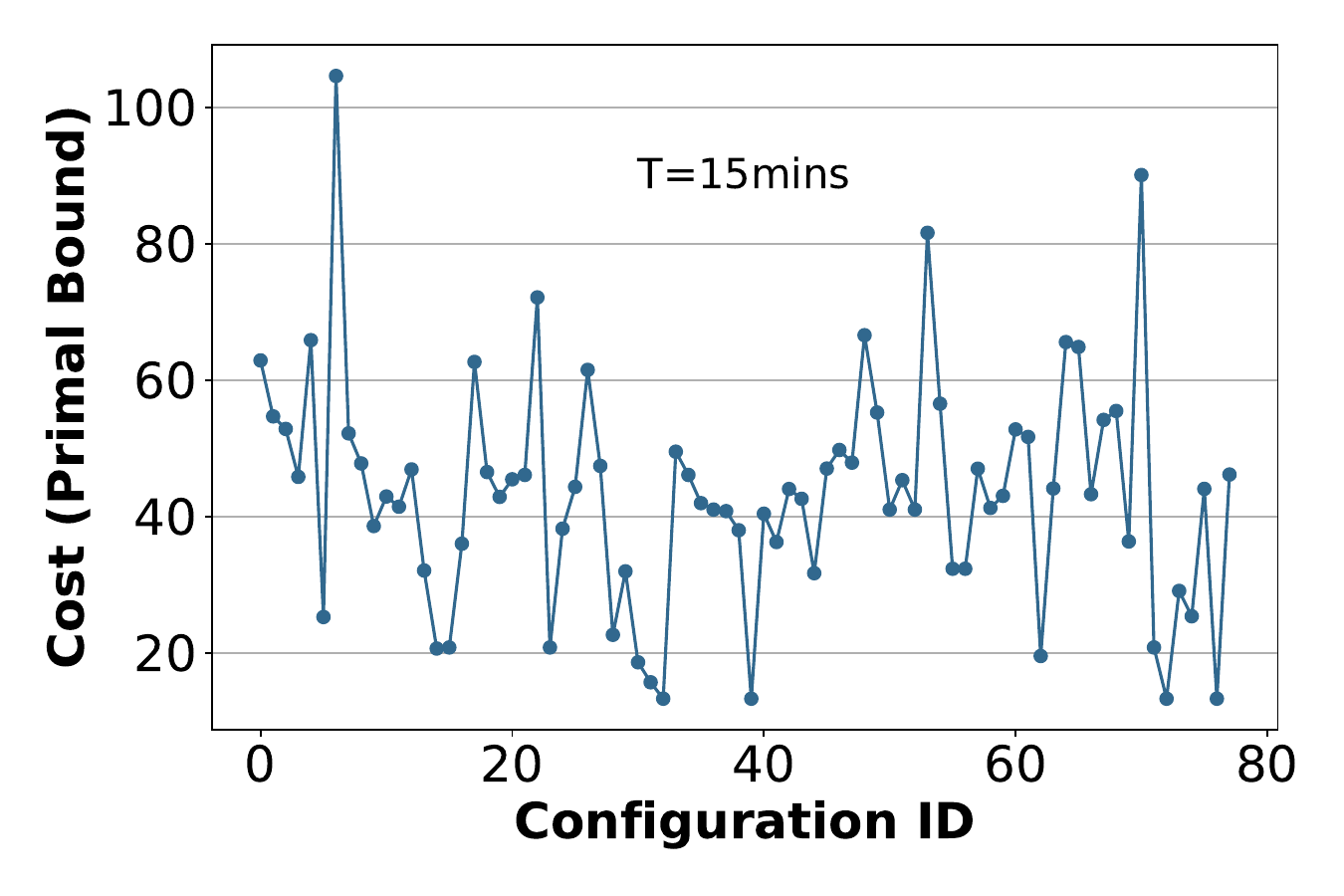}
        \centering
        \caption{Same Problem Instance}%
    \end{subfigure}%
    \begin{subfigure}{0.5\textwidth}
        \includegraphics[clip, scale=0.25]{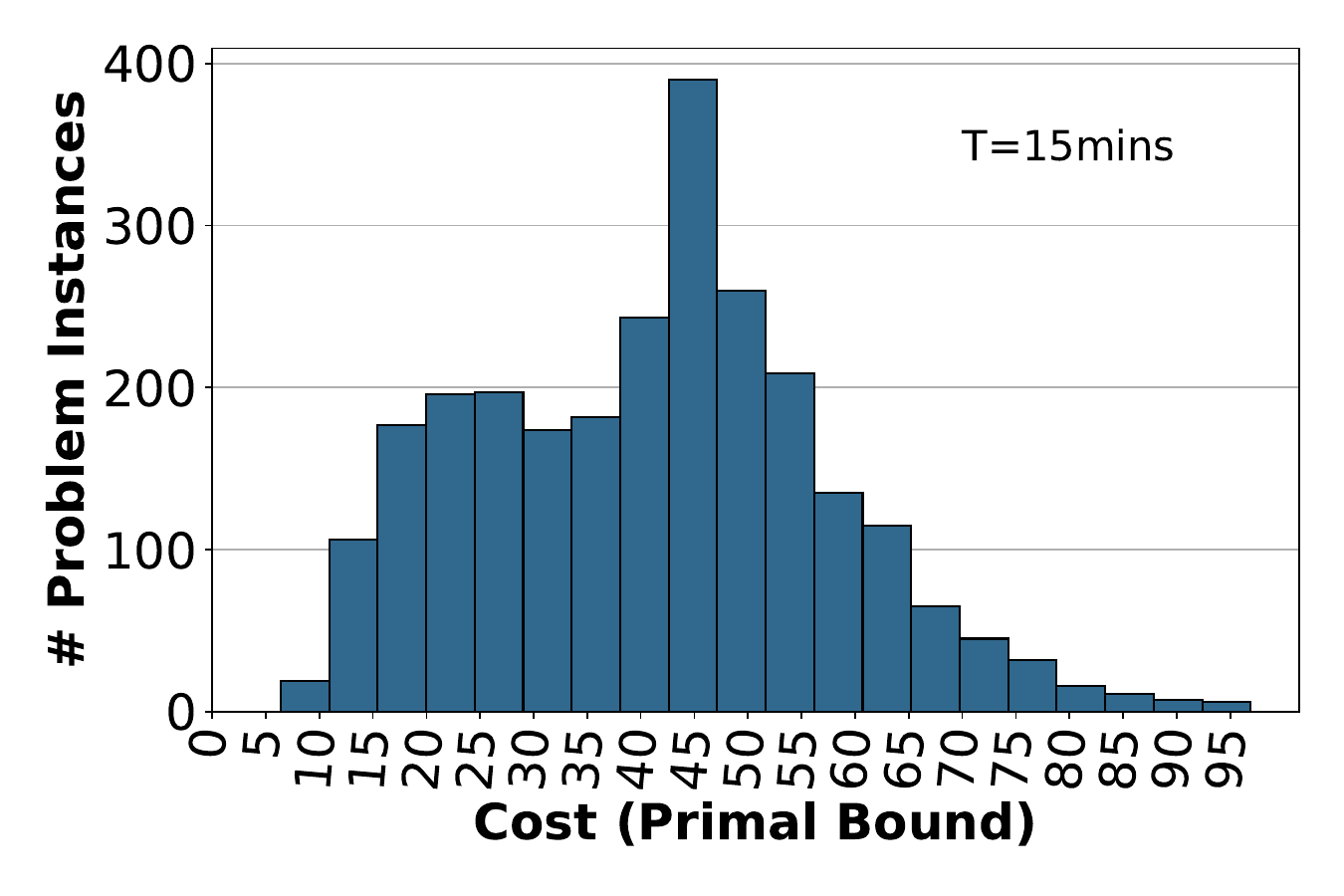}
        \centering
        \caption{Same Configuration}
    \end{subfigure}%
  \caption{Effect of configuration parameters on the solution cost using SCIP \citep{pyscipopt} ($T=15mins$ as suggested by \cite{ml4co-competition}). (a) changing the parameters of the branch-and-bound algorithm on the same instance. (b) using a single configuration on different instances. All problem instances have 195 decision variables and 1083 constraints.}
  \label{fig:config-params}
\end{figure}

\begin{figure}[t]
    \centering
    \begin{subfigure}{0.5\textwidth}
        \includegraphics[clip, scale=0.3]{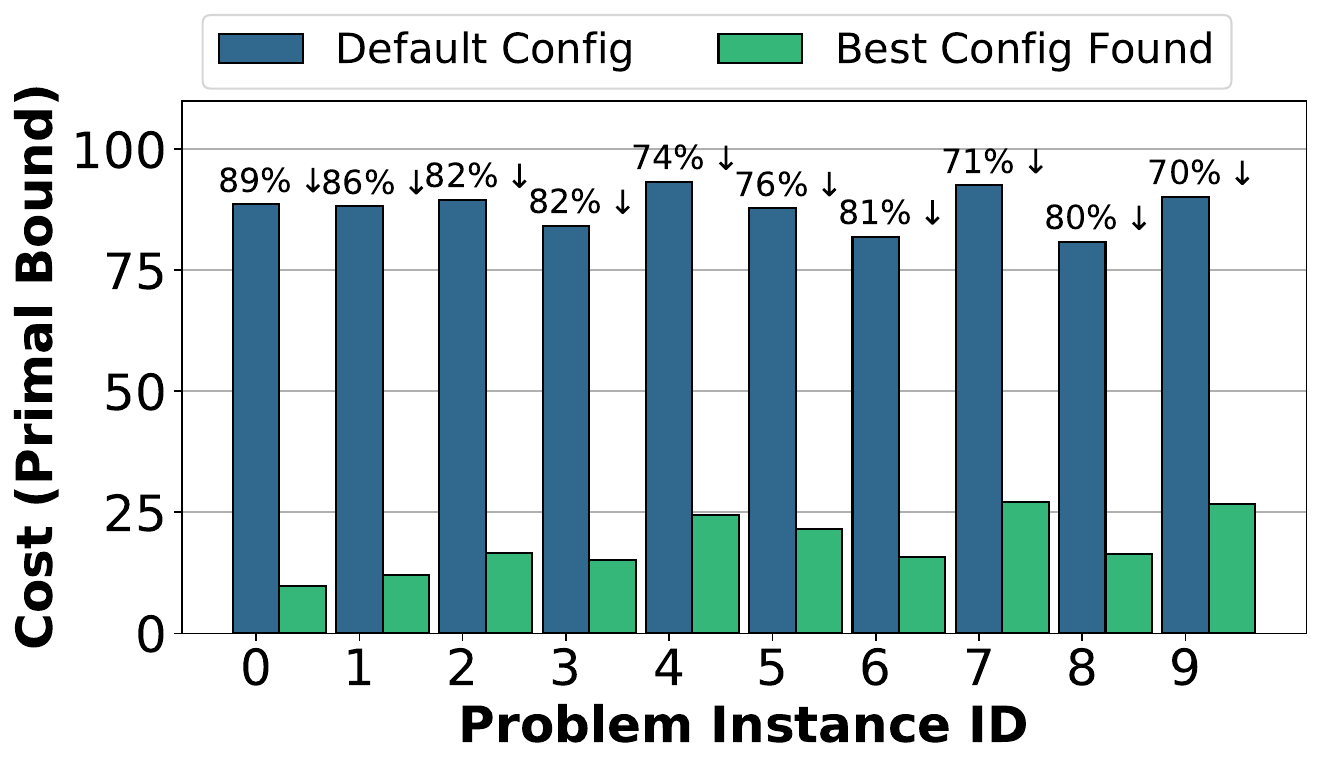}
        \centering
    \end{subfigure}
  \caption{A significant cost reduction (up to 89\%) can be achieved by searching and evaluating the configuration space of every problem instance independently. However, searching-and-evaluating takes 15 minutes (time-limit) for each evaluated configuration to complete (e.g., 10 evaluations is 150 minutes). Data shown is obtained using SMAC \cite{lindauer2022smac3} on problem instances from the ML4CO dataset \cite{ml4co-competition}.}
  \label{fig:config-params-cost-reduction}
\end{figure}

In Figure \ref{fig:config-params}, we investigate the effect of configuration parameters on problem instances from the Item Placement benchmark in the ML4CO dataset \cite{ml4co-competition}.
In Figure \ref{fig:config-params}(a), different configuration parameters directly impact the solution's cost of the same problem instance.
A solution is an assignment to the decision variables, and its cost is the value of the objective function in the formulated MILP, which is to be minimized.
In addition, using a single configuration for all problem instances does not yield the same solution's cost as shown in Figure\nobreakspace\ref{fig:config-params}(b).
As a result, branch-and-bound configuration parameters significantly affect the solution quality.
Note that increasing the time limit of the solver might not necessarily lead to better solutions since modern solvers are already heavily optimized to find solutions fast.
Moreover, time limits are usually determined by the real-world context where a solver is deployed.
Therefore, searching and evaluating configuration parameters is desirable as it can potentially improve the cost of the solutions. 
In Figure \ref{fig:config-params-cost-reduction}, we search the configuration space of every problem instance independently using SMAC \cite{lindauer2022smac3}. 
We observe that up to 89\% cost reduction can be obtained by searching for a parameters configuration that makes the branch-and-bound algorithm more efficient for the given problem instance.
Unfortunately, this search is time-consuming and cannot be performed online for every new problem instance.
Therefore, there is a need for methods to configure solvers on-the-fly while maintaining the expected cost of using a configuration tuned per instance.


Recently, machine learning (ML) has shown promising results for solving MILP problems \citep{bengio2021machine, cappart2021combinatorial}.
The motivation behind applying machine learning is to capture redundant patterns and characteristics in problems that are being solved repeatedly. 
Researchers have been able to achieve promising results by either integrating models within the solver's branch-and-bound loop\nobreakspace\citep{gasse2019exact, li2018combinatorial, wang2021bi, khalil2017learning-tree} or replacing the solver with an end-to-end algorithm that takes the raw problem instance as input and directly or iteratively output a feasible solution\nobreakspace\citep{khalil2017learning, vinyals2015pointer, bello2016neural, kool2018attention}.
Learning to configure solvers has also been explored early in \citep{kadioglu2010isac, xu2011hydra, malitsky2012instance}.
The idea is to make a solver configuration instance-specific.
In that direction, a problem instance is represented as a vector of hand-engineered features and similar instances are clustered together based on their vector representation.
Then, various sets of configurations are evaluated and assigned to each cluster.
The limitation of these works has been that features are designed rather than learned and instances within a single cluster might not, in fact, be correlated to their final solutions' costs.
Nonetheless, this direction has opened the door for instance-specific solver configuration.
More recently, meta learning on MILP has seen growing interest \citep{kruber2017learning, bonami2018learning}.
Toward learning MILP representations for solver configuration, Valentin \textit{et al.} \cite{velantin21} have proposed a supervised learning approach to predict a configuration for a specific problem instance amongst a finite set of configurations. 
However, this approach requires manually labeled data and is limited to the set of candidate configurations chosen a priori for training ($<60$). 
In other words, supervised learning restricts the ability to explore the broader configuration space once a model is trained and deployed.

\begin{figure}[t]
    \centering
    \begin{subfigure}{\textwidth}
        \includegraphics[clip, scale=0.55]{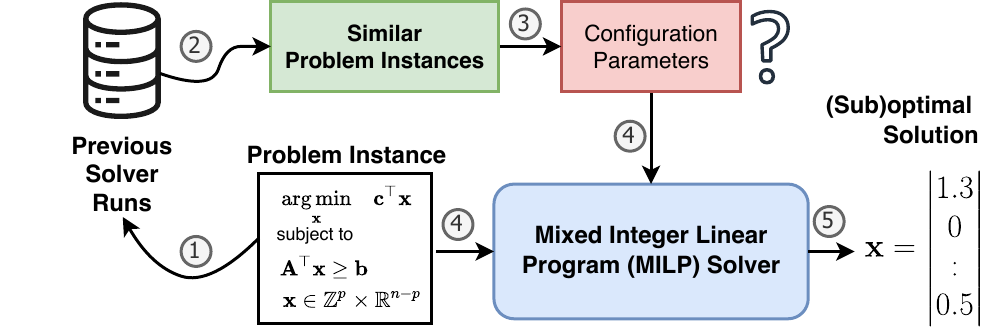}
        \centering
    \end{subfigure}
  \caption{A high-level pipeline of our proposed method. The goal is to select config. parameters based on similarity with previously solved problem instances.}
  \label{fig:high-level-overview}
\end{figure}

In this work, we address the gap in existing approaches by: (1) learning representative MILP similarities that correlate with the final solutions' costs, and (2) using a learning method that does not restrict the number of configurations to select from.  
We pursue these endeavors in a novel way through two contributions.
First, we learn an embedding space for MILP instances using \textit{Deep Metric Learning} \citep{kulis2013metric}.
Deep metric learning is a subfield of machine learning that focuses on learning a distance function between input data points using deep neural networks. The goal is to create a meaningful representation in which similar data points are mapped close together, and dissimilar data points are mapped further apart, enabling more effective clustering, classification, or retrieval tasks.
Using deep metric learning, we learn a representative embedding function for MILP, where problem instances with similar costs are closer to each other.
Unlike existing instance-aware approaches, instances' features are not hand-engineered, but learned based on \textit{Graph Convolutional Networks} \citep{kipf2016semi}, that allows our model to capture the relationships between decision variables and constraints.
Second, we predict a parameters configuration for new problem instances using nearest neighbor search on the learned metric space, which does not limit the number of configurations to predict from.
Our method is summarized in Figure \ref{fig:high-level-overview}.
The goal is to select configuration parameters based on a new problem instance's similarity with previously solved problem instances from the same distribution.
Same distribution instances are problem instances that share similar numbers of decision variables and constraints, and define a given optimization problem that is being solved repeatedly.

We show that our predictions correlate with the final solution's cost.
In other words, finding a closer instance in the learned metric space and using its well-performing configuration parameters would ultimately improve the solver's performance on the new unseen instance.
We evaluate our approach on real-world benchmarks from the ML4CO competition dataset \citep{ml4co-competition} using SCIP solver \citep{gamrath2020scip}, and compare against both using an incumbent configuration from SMAC \citep{lindauer2022smac3}, and predicted configurations from existing instance-aware methods.
Our method solves more instances with lower costs than the baselines and achieves up to 38\% improvement in the cost.

%% file: 2_related_work.tex
\textbf{Machine Learning for Combinatorial Optimization.}
Learning-based optimization methods have seen growing interest lately \citep{bengio2021machine, cappart2021combinatorial}.
Broadly speaking, they can be divided into methods inside the solvers \citep{khalil2017learning-tree, gasse2019exact, li2018combinatorial, wang2021bi}, methods outside the solvers \citep{kruber2017learning, bonami2018learning}, and methods that replace the solvers \citep{khalil2017learning, vinyals2015pointer, bello2016neural, kool2018attention}. 
Our work is amongst methods outside the solver, which aims at improving the solver's performance by instantly predicting instance-aware parameters configuration.
This is orthogonal to existing work and can benefit from existing hyper-parameter search methods when performed offline.

\noindent \textbf{Instance-aware Solver Configuration.}
Instance-aware configuration methods have been explored early in ISAC \citep{kadioglu2010isac}, which stands for Instance-Specific Algorithm Configuration.
The method extracts features from problem instances and assigns problem instances with similar feature vectors to a cluster using g-means clustering.
Features include problem size, proportion of different variable types (e.g., discrete vs continuous), constraint types, coefficients of the objective function, the linear constraint matrix and the right hand side of the constraints.
After that, assuming that problem instances with similar features behave similarly under the same configuration, local search is used to find good parameters for each cluster of instances.
Although this approach allows us to bypass the expensive search-and-evaluate at deployment time, features are hand-engineered and need to be adapted for each problem, e.g., as in \citep{ansotegui2016maxsat}.
In other words, the algorithm requires further refining of the distance metric in the feature space so that it can find better clusters.
Hydra-MIP \cite{xu2011hydra} enhanced this approach by including features from short solver runs before selecting a configuration for a complete solver run.
It also uses pair-wise random forests to select amongst candidate configuration parameters.
Our approach is different since problem features are \textit{learned} during training, and correlates similarity to the costs of final solutions.
Moreover, these approaches assign a single parameters configuration for each cluster, which limits the portfolio of configurations available at inference time.
More recently, supervised deep learning was investigated in \cite{velantin21}.
The method selects a limited number of configuration parameter sets, and collects training data by running the solver using the selected configurations on all problem instances separately.
Using the labeled data, it can predict the cost of running the solver on a new unseen instance using one of the configurations used during training.
Aside from the massive labeled data required for training, this approach limits the potential of exploring other sets of configurations after the model is trained and deployed.
Exploring further solver configurations would require solving and labeling more problem instances, then re-training the model.

\noindent \textbf{Shallow Embedding vs. Deep Embedding.}
Instance-aware configuration methods represent a MILP instance as a vector of values that encapsulate the primary characteristics of the problem instance.
The objective of a specific embedding (i.e., encoding) method is to ensure that similarity in the embedding space (e.g., dot product) closely mirrors the similarity found in the original problem representation.
The effectiveness of an embedding method is determined by its ability to uniquely identify and distinguish between problem instances that may share similar properties (such as the number of decision variables and constraints) but exhibit differences in their solutions' costs within the same solving environment.
A powerful embedding method can accurately differentiate between such instances, enabling more effective and tailored configuration selection for each problem instance.
\textit{Shallow} embedding is the simplest encoding approach, where the encoder is just an embedding lookup.
For example, in ISAC \cite{kadioglu2010isac}, problem instances are encoded as feature vectors for the Set Covering problem that include a normalized cost vector $\mathbf{c}$, bag densities, item costs and coverings, in addition to other density functions.
These values are aggregated (using minimum, maximum, average and standard deviation) to construct the final feature embedding of the problem instance.
Similarly, authors in \citep{ansotegui2016maxsat} use ISAC's method and focuses on the maximum satisfiability problem (MaxSAT), with hand-engineered features that include problem size, balance features and local search features.
Hydra-MIP \citep{xu2011hydra} extracts more features by executing short runs of the solver (CPLEX) using a default configuration on each new instance.
These features include pre-solving statistics, cutting planes usage, and the branch-and-bound tree information.
While shallow embedding is straightforward to compute, these encoders are non-injective\nobreakspace\citep{xie2016unsupervised}.
That is, different MILP instances could have the same embedding using a shallow feature vector.
Moreover, and by design, shallow embeddings are not necessarily correlated with the final costs of the solver's solutions.

In contrast, \textit{Deep} encoders learn the embedding function during training according to a defined loss function.
In other words, a deep encoder is characterized by learnable parameters of a deep neural network that defines embedding similarity based on a loss function.
In our method, we train a Graph Convolutional Network (GCN) to embed MILP instances to an embedding space where the similarity of instances is defined based on their final solutions' costs in the same solving environment.
Deep embedding is injective and uniquely encodes problem instances even if they have the same shallow embedding (e.g., number of decision variables).
In Section \ref{sec:expr}, we show that when the size of the problem remains relatively similar, but the coefficients or structure vary, deep embedding has a larger discriminative power over shallow embedding.
In problems where the problem size varies significantly, shallow embedding could be enough.
In designing a system that is invariant to the problem size, deep embedding addresses the need without hand-engineering features for each problem separately.

\noindent\textbf{Configuration Space Search.}
During the process of identifying similar problem instances, addressing the selection of parameter configurations remains a challenge. 
In situations where ample time is available for exploration, such as testing different parameter configurations on a single problem instance, several methods have been proposed to navigate this vast search space. 
These methods aim to identify a single robust configuration\footnote{Also called incumbent configuration in the context of parameters configuration search; not to be confused with the incumbent solution of the solver itself, which is the $\mathbf{x}$'s assignment with minimum cost of the MILP objective function.} across a collection of problem instances, denoted as $\mathcal{I}$.
Random search \citep{bergstra2012random}, evolutionary algorithms \citep{olson2016evaluation}, Bandit methods \citep{li2017hyperband}, and Bayesian-based optimization \citep{shahriari2015taking} are among the top-performing methods that have been applied successfully in various optimization contexts \citep{lopez2016irace, birattari2009tuning, birattari2002racing, maron1997racing, hoos2012automated, kerschke2019automated}.

The SMAC package \citep{lindauer2022smac3} is an instance of model-based optimization that employs Bayesian optimization for searching parameters configuration \citep{hutter2011sequential}.
The central concept of SMAC revolves around building a probabilistic model, specifically a random forest model, which predicts the performance of an algorithm on a set of instances, given a specific configuration.
By sequentially updating this model based on the observed performance of algorithm configurations, SMAC is able to efficiently search for an optimal or near-optimal configuration within a pre-defined search space.
The key components of SMAC include the acquisition function \citep{jones1998efficient}, which guides the search process in terms of exploration and exploitation trade-off, and the intensification procedure \citep{li2017hyperband}, responsible for selecting a new incumbent configuration.
The most common acquisition function used in SMAC is the Expected Improvement (EI) function \citep{snoek2012practical, hutter2010time}, which aims to minimize the expected runtime of the target algorithm.

Another popular package for algorithm configuration, the irace package \cite{lopez2016irace} is based on the Iterated Racing framework, which is a derivative of the F-race procedure \cite{birattari2010f, balaprakash2007improvement}.
The main idea behind irace is to iteratively sample and compare algorithm configurations on an increasing set of problem instances, using statistical tests to eliminate poorly performing candidates.
This iterative process continues until a termination criterion is met, usually when a maximum number of iterations or a maximum time is reached.
The irace package is particularly well-suited for discrete and categorical parameter spaces, as it does not require any explicit modeling of the performance landscape.
The search process is guided by a combination of adaptive sampling and statistical tests, which provide a balance between exploration and exploitation.
The elimination of underperforming configurations is carried out using a statistical test, most commonly the Friedman test or the two-sample t-test, which considers the performance of the remaining configurations.
The key differences between both packages is that SMAC adopts a model-based optimization strategy with Bayesian optimization, building a surrogate model to predict algorithm performance, while irace is a model-free approach relying on iterative sampling and statistical tests to identify the best-performing configurations.

In this paper, our method of selecting parameters configurations based on problem instance similarity is agnostic to the package used for the offline configuration search phase.
We use SMAC for its interoperability with the SCIP solver \citep{gamrath2020scip} through its Python binding \citep{pyscipopt} along with the PyTorch Ecosystem \citep{musgrave2020pytorch} used for training the deep metric learning model.
Nonetheless, after the model training phase is completed, which entails learning the similarity, the system illustrated in Figure \ref{fig:milptune-overview} can be adapted to incorporate the irace package for an additional offline search of the configuration space of previously solved problem instances. 
Our approach eliminates the necessity to retrain the previously acquired similarity models.

\noindent\textbf{Predicting Solver Configuration.}
Aside from using shallow or deep embedding, predicting a solver configuration for an unseen instance requires selecting an already-evaluated configuration from similar instances in the embedding space.
ISAC \cite{kadioglu2010isac} and MaxSAT \cite{ansotegui2016maxsat} use G-means clustering to cluster similar instances.
Then, they assign a single configuration to each cluster to be used for new instances that are embedded into that cluster.
This approach evolves by refining the distance metric in the feature space so that it can find better clusters in future iterations.
Hydra-MIP \cite{xu2011hydra} uses pair-wise weighted random forests (RFs) to select amongst $m$ algorithms for solving the instance, by building $ m (m - 1) / 2 $ RFs and taking a weighted
vote.
When the number of parameters configuration to select from is large (i.e., large $m$), calculating pair-wise RFs becomes computationally infeasible.
In our method, we use k-nearest neighbor (KNN) to predict a parameter configuration from the closest problem instance in the learned embedding space.
This allows our approach to scale the exploration of configuration parameters without the restriction of refining clusters, or re-building a limited number of pair-wise RFs.
In both Hydra-MIP\nobreakspace\cite{xu2011hydra} and our method, $k$ configurations can be predicted at once (with ranks) to potentially run the solver using multiple configurations in parallel.
Table \ref{tab:summary-shallow-vs-deep} summarizes the differences between our method and existing instance-aware solver configuration methods.

\begin{table}[t]
\renewcommand{\tabcolsep}{2pt}
\caption{Summary of instance-aware solver configuration methods. Details of the methods are described in Section \ref{sec:related-work}.}
  \centering
  \begin{tabular}{llll}
    \toprule
     & \textbf{ISAC} \citep{kadioglu2010isac, ansotegui2016maxsat}  & \textbf{Hydra-MIP} \citep{xu2011hydra} & \textbf{Our Method}  \\
     \midrule
     \textbf{Features} & Hand-crafted & Hand-crafted & Learned \\
     \textbf{Embedding} & Shallow & Shallow & Deep \\
     \textbf{Injectivity} & Non-injective & Non-injective & Injective \\
     \textbf{Offline Search} & Genetic Algorithm & Regression/Iterative & Bayesian Search \\
     \textbf{Inference} & G-means Clustering & Random Forests & KNN \\
     \textbf{\#Configs Predicted} & 1 & k (hyperparameter) & k (hyperparameter) \\
    \bottomrule
  \end{tabular}
  \label{tab:summary-shallow-vs-deep}
\end{table}

%% file: 3_prelim.tex
\subsection{MILP Formulation}
In this work, we consider MILP instances formulated as:

\begin{equation}
	\underset{\mathbf{x}}{\operatorname{arg\,min}} \quad \mathbf{c}^\top\mathbf{x} \text{,} \qquad
	\text{subject to} \quad \mathbf{A}^\top\mathbf{x}  \geq  \mathbf{b} \text{,}
	\qquad
	\text{and} \quad \mathbf{x} \in \mathbb{Z}^p \times \mathbb{R}^{n-p}
    \label{eq:problem-formulation}
\end{equation}
where $\mathbf{c} \in \mathbb{R}^n$ denotes the coefficients of the linear objective, 
$\mathbf{A} \in \mathbb{R}^{m \times n}$ and $\mathbf{b} \in \mathbb{R}^m$ denote  the coefficients and upper bounds of the linear constraints, respectively.
$n$ is the total number of decision variables, $p \leq n$ is the number of integer-constrained variables, and $m$ is the number of linear constraints.
The goal is to find feasible assignments for $\mathbf{x}$ that minimize the objective $\mathbf{c}^\top\mathbf{x}$.
A MILP solver constructs a search tree to find feasible solutions with minimum costs.
The cost of the solution found by the solver by the end of its search, or if the time limit is reached, is called the primal bound. 
It serves as an upper bound to the set of feasible solutions.
While there are other methods to measure the solver's performance, (e.g., dual bound, primal-dual gap, primal-dual integral \citep{achterberg2007constraint}), we adopt the primal bound at the end of the time limit for the purpose of training the metric learning model.

\subsection{Graph Neural Networks}

Graph Neural Networks (GNNs) offer a powerful paradigm for analyzing complex relational data, which is often encountered in Operations Research problems \citep{wang2023dynamic, lee2023efficient}.
GNNs are designed to learn meaningful representations of nodes in a graph by incorporating both node features, edge features and graph structure.
The core principle behind GNNs is message-passing, where information is aggregated from neighboring nodes to update the representations iteratively.

The message-passing framework for GNNs can be formalized as follows. Let $G = (V, E)$ denote a graph with nodes $V$ and edges $E$.
Each node $v_i \in V$ is associated with a feature vector $x_i$.
The goal is to learn a representation $h_i$ for each node $v_i$.
The message-passing process in GNNs typically consists of $L$ layers, where each layer $l$ updates the node representations based on the previous layer's representations. The update at each layer can be expressed as:

\begin{align}
m_{i}^{(l)} &= \sum_{j \in \mathcal{N}(v_i)} M^{(l)}(h_{j}^{(l-1)}, h_{i}^{(l-1)} e_{ji}), \\
h_{i}^{(l)} &= U^{(l)}(h_{i}^{(l-1)}, m_{i}^{(l)})
\end{align}

where $\mathcal{N}(v_i)$ denotes the set of neighboring nodes of $v_i$, $M^{(l)}$ is a message function that computes the messages $m_{i}^{(l)}$ to be sent from node $j$ to node $i$ at layer $l$, $U^{(l)}$ is an update function that computes the new node representation $h_{i}^{(l)}$ using the aggregated messages, and $e_{ji}$ represents the edge features, if present.
Both $M^{(l)}$ and $U^{(l)}$ are typically implemented as neural networks, allowing GNNs to learn complex, nonlinear relationships between nodes.
In this work, we use a GNN from \citep{gasse2019exact} to model the relationships between decision variables and constraints in Equation \ref{eq:problem-formulation}.

\vspace{0.2in}
\subsection{Metric Learning}
Deep learning models require a vast amount of data in order to make reliable predictions.
In a supervised learning setting, the goal is to map inputs to labels as in a standard classification or regression problem.
When the number of classes is huge, supervised learning fails to address real-world applications.
For example, face verification systems have a large number of classes, but the number of examples per class is small or non-existent \citep{schroff2015facenet}.
In this case, the goal is to develop a model that learns object categories from a few training examples.
But deep learning models do not work well with a small number of data points.
In order to address this issue, we learn a similarity function between data points, which helps us to predict object categories given small data for training.
This paradigm is known as metric learning \citep{kulis2013metric}.
In this paradigm, a model is trained to learn a distance function (or similarity function) over the inputs themselves.
Here, similarity is subjective, so the distance may have a different meaning depending on the data. 
In other words, the model learns relationships in the training data regardless of what it actually means in its application domain. 
Metric learning has seen growing adoption in real-world applications, such as face verification \citep{schroff2015facenet, wang2018cosface, deng2019arcface}, video understanding \citep{lee2018collaborative} and text analysis \citep{davis2008structured}.

Measuring distances is a critical aspect of metric learning. 
Given two instances of some object representation, $\mathcal{I}_i$ and $\mathcal{I}_j$, a distance function, $d$, measures how far the two instances are from each other.
The Euclidean distance is challenging to reason about in higher dimensions even if the data is perfectly isotropic and features are independent from each other.
Therefore, the goal is to define new distance metrics in higher dimensional spaces that are based on the properties of the data itself.
These are non-isotropic distances reflecting some intrinsic structures of the data. 
A parametric model is trained to project instances to the new metric space through either a linear transformation of the data such as the Mahalanobis distance \citep{de2000mahalanobis}, or a non-linear transformation of the data using deep learning \citep{kaya2019deep}.
This projection step allows the Euclidean distance to capture relationships between the features that are non-linear or more complex; in our case, the correlation between a problem instance structure and its solution's cost on a given solving environment.

In metric learning, instead of requiring labels for training, the model requires weak supervision at the instance level, where triplets of (anchor $a$, positive $p$, negative $n$) are fed into the model. 
The model is trained to learn a distance metric that puts positive instances close to the anchor and negative instances far from the anchor.
This is achieved by a Triplet loss function \citep{schroff2015facenet}:

\begin{equation}
    L = \sum_{i}^{N} [\lvert\lvert f(\mathcal{I}_i^a) - f(\mathcal{I}_i^p) \lvert\lvert^2 - \lvert\lvert f(\mathcal{I}_i^a) - f(\mathcal{I}_i^n) \lvert\lvert^2 + \alpha]_+
    \label{eq:loss-function}
\end{equation}
where $N$ is the number of triplets sampled during training.
$\mathcal{I}^a$, $\mathcal{I}^p$ and $\mathcal{I}^n$ represent the anchor instance, similar instance and dissimilar instance, respectively.
$f$ is a parametric model that projects instances to a learned metric space.
The loss increases when the first squared distance (anchor-positive) is larger than the second squared distance (anchor-negative).
So, $f$ is trained to decrease this loss.
In other words, it tries to make the first squared distance smaller, and the second square distance larger.
Here, the loss, $L$, will be equal to zero if the first squared distance is $\alpha$-less than the second squared distance.
While there are other variants of the loss functions for metric learning, e.g. Contrastive Loss \citep{koch2015siamese}, Triplet loss can provide more stable training compared to contrastive loss, as it considers both positive and negative examples simultaneously for each anchor point \citep{wang2021understanding}.
In addition, Triplet loss aims to ensure that the distance between the anchor and positive example is smaller than the distance between the anchor and negative example, by a margin.
This allows the model to learn a more balanced similarity metric, especially for complex structures such as MILP.
In Section \ref{sec:methodology}, we present a number of modifications during training in order to avoid having a zero-loss early during training.

%% file: 4_motivation.tex
\begin{figure}[t]
    \centering
    \begin{subfigure}{0.5\textwidth}
        \includegraphics[clip, scale=0.3]{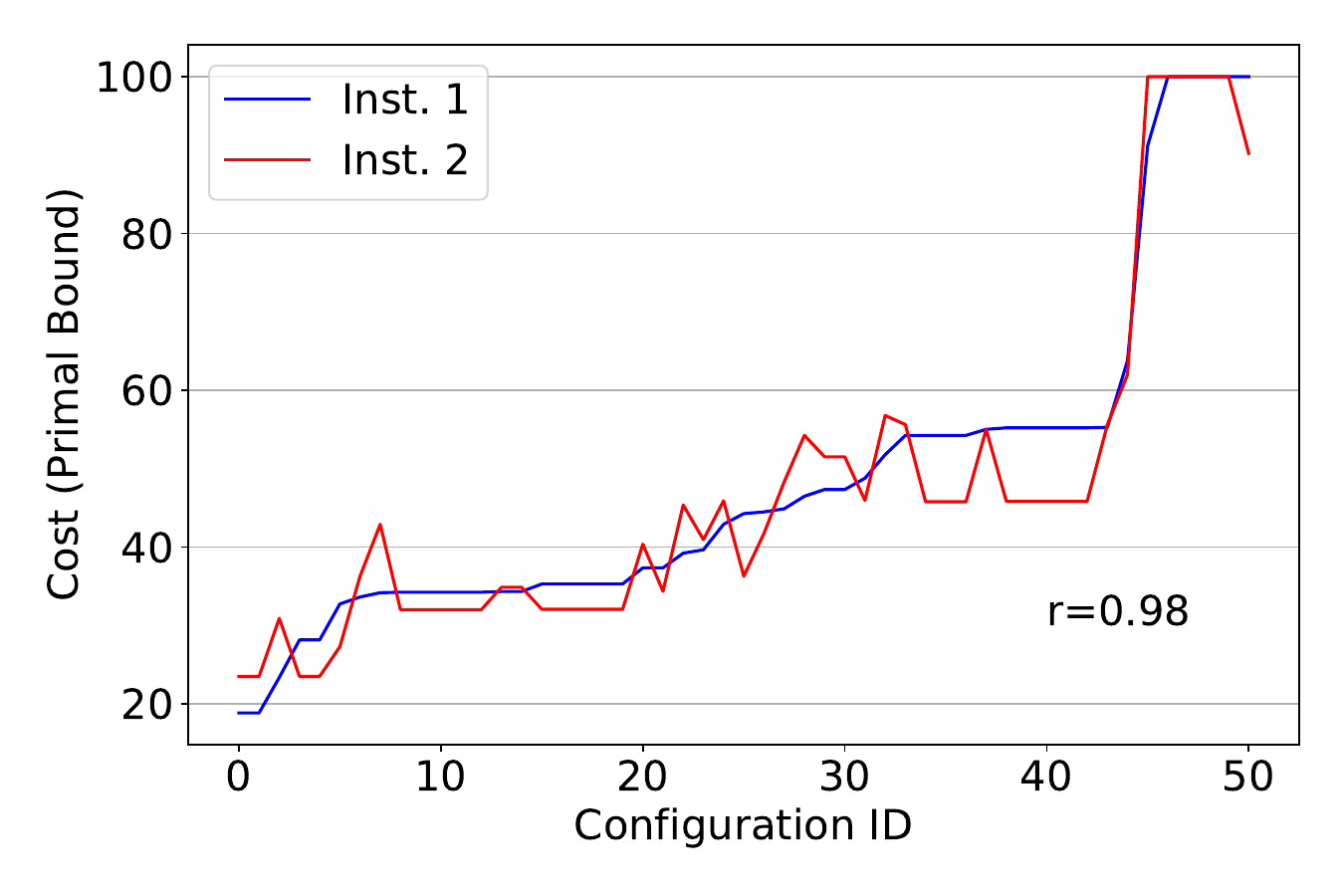}
        \centering
    \end{subfigure}
  \caption{Two MILP instances that have similar costs when using the same default configuration have similar costs when using other configurations. Other environment variables are fixed: solver version, machine (cpu and memory) and random seed. $r$ is the Pearson correlation coefficient.}
  \label{fig:two-inst-correlation}
\end{figure}

The fundamental motivation of our work is to define similarity among MILP instances based on their final solutions' costs after running the solver in the same environment (i.e., host machine, software environment, configuration parameters, time limit, and random seed), and under the assumption that all MILP instances are coming from the same problem distribution.
Same distribution instances are problem instances that share similar number of decision variables and constraints, and define a problem that is being solved repeatedly.
To our knowledge, no prior work has explored correlating MILP similarity to the costs (objective function) of their final solutions.

\begin{figure}[t!]
    \centering
    \begin{subfigure}{0.5\textwidth}
        \centering
        \includegraphics[clip, scale=0.25]{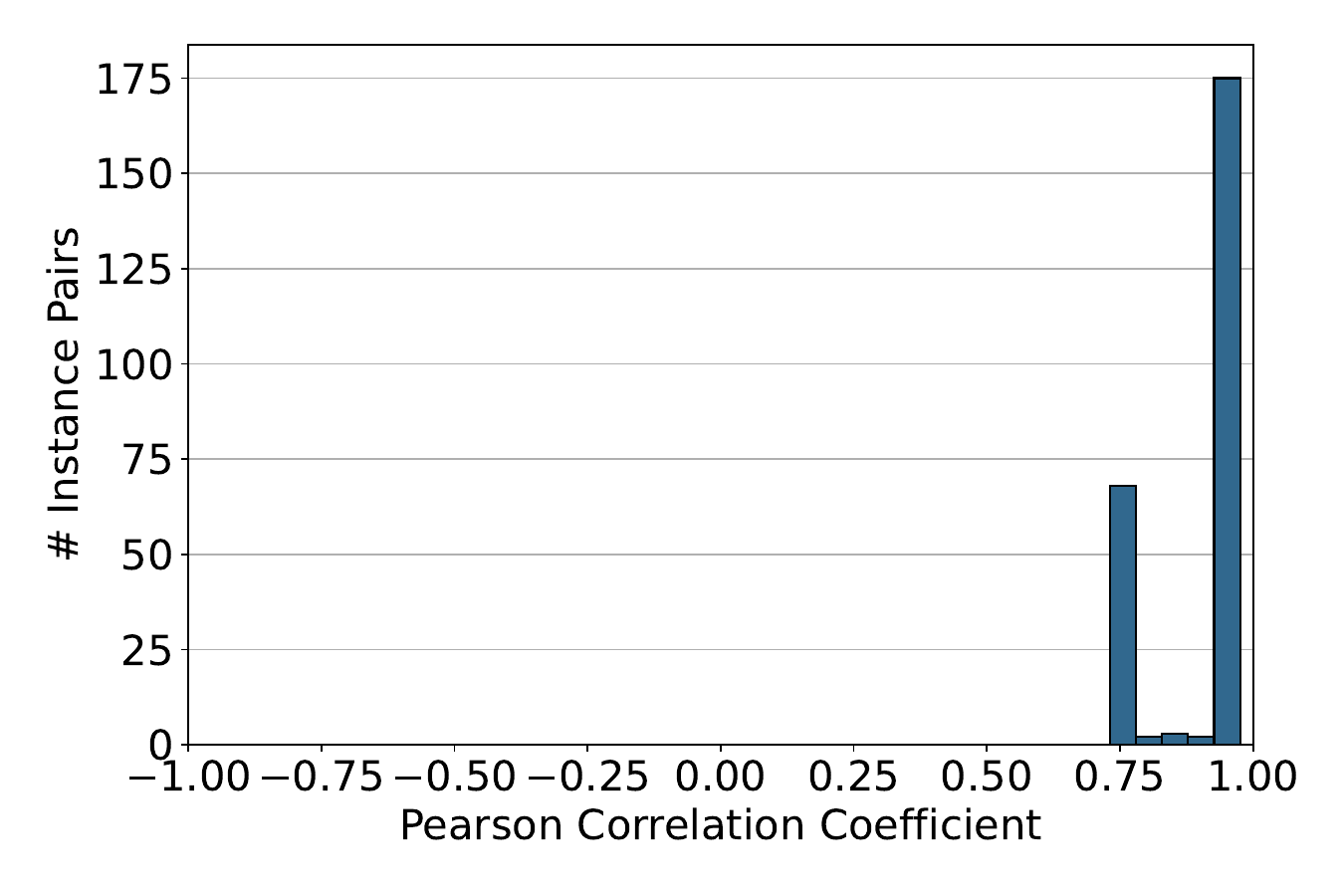}%
        \caption{\centering Similar instance pairs $\lvert C(\mathcal{I}_i) - C(\mathcal{I}_j) \rvert \leq C_{thr}$}%
    \end{subfigure}%
    \begin{subfigure}{0.5\textwidth}
        \centering
        \includegraphics[clip, scale=0.25]{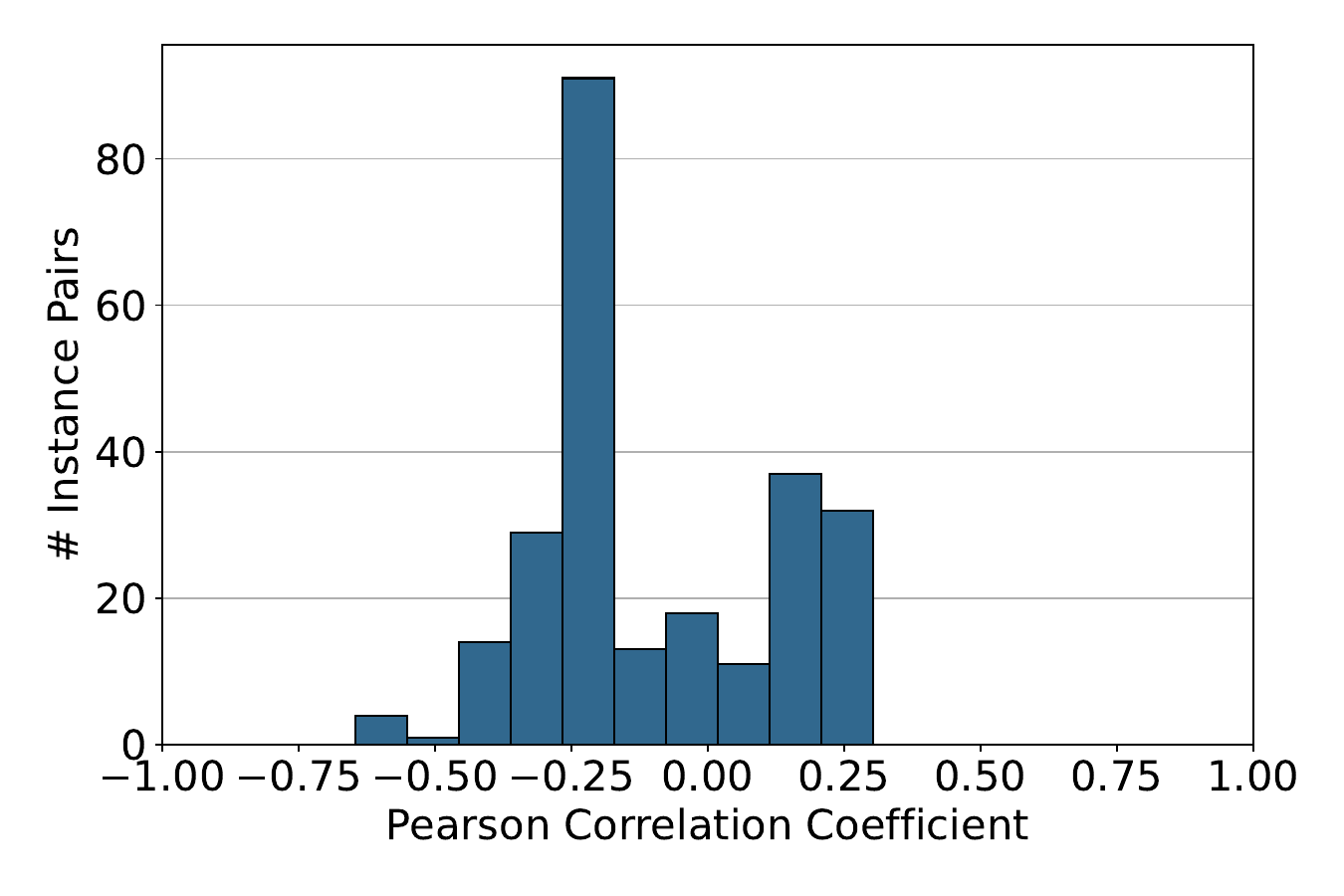}
        \caption{\centering Dissimilar instances pairs $\lvert C(\mathcal{I}_i) - C(\mathcal{I}_j) \rvert \gg C_{thr}$}%
    \end{subfigure}%
  \caption{Correlation between problem instances and their costs at different parameters configurations. (a) Pairs of instances with similar costs when using the default configuration are solved simultaneously using other configurations and their cost correlations are measured using Pearson correlation coefficient. (b) Likewise, pairs of instances with large cost difference when using the default configuration are solved simultaneously using other configurations and their cost correlations are measured. Problem instance pairs were retrieved from the Item Placement dataset \cite{ml4co-competition} where each instance has 185 decision variables and 1083 constraints. The solver is run on 250 random similar pairs and 250 random dissimilar pairs using eight different configurations for each run. Here, $C_{thr} = 1$, and the $C$ $\in [0, 100]$.}
  \label{fig:correlation}
\end{figure}

First, we validate the assumption that MILP instances which have similar costs when solved in a specific environment would have similar costs when changing the solver configuration. 
For this validation, we use the Item Placement benchmark from the ML4CO dataset \cite{ml4co-competition}.
We run the solver on 25\% of the training dataset (i.e., 2500 MILP instances) using the default solver configuration and a time limit of 15 minutes.
Each solver run is executed independently and is given the same compute and memory resources.
In Figure \ref{fig:two-inst-correlation}, we select two MILP instances that have similar costs ($\lvert C(\mathcal{I}_1) - C(\mathcal{I}_2) \rvert \leq C_{thr}$) when using the default solver configuration, and solve both of them independently using different sets of configurations while still fixing all other hyper-parameters (i.e., cpu, memory, solver version, time limit and random seed).
Here, $C_{thr} = 1$, and the $C$ $\in [0, 100]$.
We observe that the costs of the two instances are indeed positively correlated with a Pearson correlation coefficient of $r=0.98$.
We extend this investigation to validate if this is the case for other pairs of similar and dissimilar MILP instances in the collected dataset.
So, we select 250 MILP instances (10\%) that are similar in their costs, and another 250 MILP instances (10\%) that are largely dissimilar in their costs. 
We run each pair of instances independently using eight other solver configurations and report their final solutions' costs.
Figure \ref{fig:correlation} shows a histogram of the Pearson correlation coefficient for similar and dissimilar pairs of instances.
We observe that similar pairs of instances have a Pearson correlation coefficient $\gt 0.75$, which indicates a high positive correlation, while dissimilar pairs of instances either have a small correlation coefficient $\lt 0.3$, or a negative coefficient indicating an inverse correlation.
This finding confirms that if we are able to define MILP similarity based on their final solutions' costs (unlike \cite{kadioglu2010isac, xu2011hydra} that define similarity based on hand-crafted features without correlation to the final solutions' costs), we will be able to predict an effective parameters configuration for the solver for a new unseen MILP instance by fetching a previously-evaluated configuration from a similar instance.

\vspace{0.2in}

%% file: 4_method.tex
\begin{figure}[t]
  \centering
  \includegraphics[clip, trim=2.3cm 0cm 0.8cm 0cm, scale=0.465]{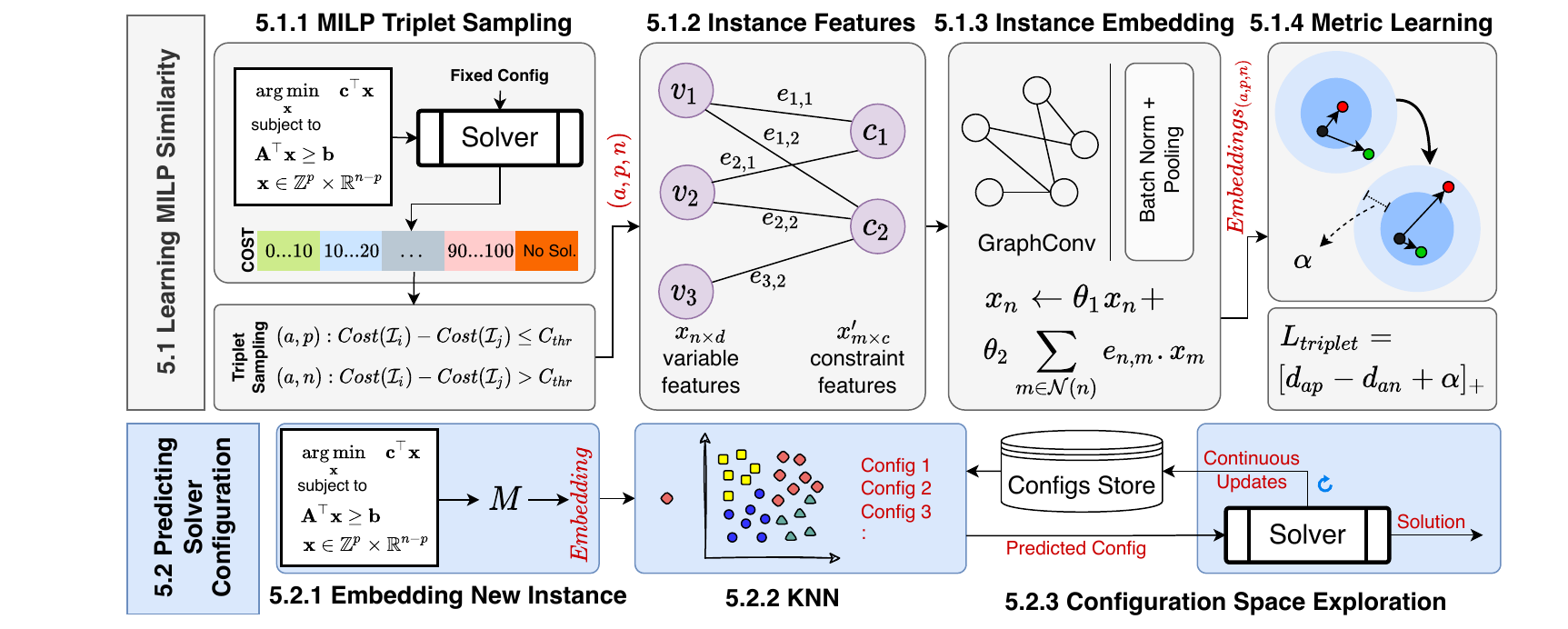}
  \caption{Overview of our method. Triplet samples are first collected on a few instances using the default solver configuration. Instance features are extracted as a bipartite graph \citep{gasse2019exact}, then embedded using a graph convolutional network. A triplet loss \citep{schroff2015facenet} function is used to train the model end-to-end. $C_{thr}$: cost threshold for similarity, $a$: anchor instance, $p$: positive/similar instance, $n$: negative/dissimilar instance, $\theta$: learnable parameters of the GNN, $d$: distance between embeddings, as defined in Equation \ref{eq:loss-function}. $M$ refers to the trained GNN model. KNN refers to using k-nearest neighbors to find a similar instance in the learned embedding space. The Config Store is a continuously-updated data store to expand the portfolio of explored configurations after model deployment.}
  \label{fig:milptune-overview}
\end{figure}

In order to define MILP similarity based on the final solutions' costs, our approach is to use \textit{Deep Metric Learning} to learn the instance embeddings, and based on that predict instance-aware parameters configuration.
Figure \ref{fig:milptune-overview} shows an overview of our methodology.
In contrast to supervised learning where a large amount of data needs to be collected in order to train the model, we collect training data on a small subset of the problem instances available.
The method is divided into two major parts: (1) a training phase to learn MILP similarities based on costs, and (2) an inference phase to predict a parameters configuration for a new MILP instance.
We present the details for each phase in Subsections \ref{sec:sub:triplet-sampling} and \ref{sec:sub:predicting-configuration}, respectively.

\subsection{Learning MILP Similarity}
\label{sec:sub:triplet-sampling}
In the training phase (5.1) of Figure \ref{fig:milptune-overview}, and given two MILP instances, $\mathcal{I}_i$ and $\mathcal{I}_j$, the goal is to train a parametric model that recognizes whether $\mathcal{I}_i$ and $\mathcal{I}_j$ are similar or not.
As discussed in Section \ref{sec:preliminaries}, similarity is subjective and depends on the domain.
In our case, there is no natural way to find out whether two instances are similar or not just from their given problem formulation (Equation \ref{eq:problem-formulation}).
Even though one could map it to a graph isomorphism problem, small perturbations of $\mathbf{A}$ can lead to different solutions from the solver.
For example, a slight change in a constraint's coefficients could make the constraint trivial, or make the MILP instance infeasible \cite{velantin21}.

We divide the training stage into four main steps.
In the first step (5.1.1), we sample MILP instances from the training set based on their final solutions costs.
The steps in the middle (5.1.2 and 5.1.3) include a Graph Convolutional Network (GCN) model that extracts features from problem instances and passes them through convolutional layers of learnable parameters that reduce the loss during training.
In the last step (5.1.4), we define our loss function with the goal of bringing the learned embeddings of similar instances closer to each other, and dissimilar instances further from each other.

\subsubsection{MILP Triplet Sampling}

In our method, if the difference between the solution cost of instance $\mathcal{I}_i$ and $\mathcal{I}_j$ is below a certain threshold C$_{thr}$, then $\mathcal{I}_i$ and $\mathcal{I}_j$ are considered similar for the purpose of training the model.
If the cost difference is above C$_{thr}$, the instances are considered dissimilar.
In the triplet sampling step, the goal is to look up for similar and dissimilar instances in the training dataset.
Algorithm\nobreakspace\ref{algo1} shows the steps for the mining and training procedures.
In line 1, we sample an arbitrary anchor instance ($a$).
In line 2, we sample a similar instance ($p$), in which the difference in their costs under the default solver configuration is less than a threshold.
In our work, we introduce a new sampling schedule for the training procedure.
The goal is to avoid crunching the loss (Equation \ref{eq:loss-function}) to zero prematurely.
Therefore, in line 3, we start with hard negative sampling by looking for instances that have a cost difference much larger than the threshold.
The idea is that when starting with these negative pairs ($a$, $n$), the model gets a chance to to push their embeddings further away from each other when training for a certain number of epochs (line 4).
Then, in lines 5-6, this restriction is relaxed and the training loop starts seeing negative instances that have slightly larger cost difference than positive instances.
Theoretically, triplet sampling can be done using any other defined measure of similarity.
We chose to use the cost after running the solver in order to correlate similarity with the final solutions costs.

\begin{algorithm}[t!]
\caption{MILP Triplet Sampling}\label{algo1}
\textbf{Input:} Training dataset of MILP instances ($\mathcal{I}$) \\
\textbf{Input:} Costs using default solver configuration ($C_i$) \\
\textbf{Input:} Cost threshold ($C_{thr}$) \\
\textbf{Output:} Triplets ($a$, $p$, $n$)
\begin{algorithmic}[1]
\State Select an arbitrary anchor instance ($a$) from $\mathcal{I}$.
\State Find a similar problem instance ($p$) such that $\lvert C(\mathcal{I}_a) - C(\mathcal{I}_p) \rvert < C_{thr}$
\State Find a dissimilar problem instance ($n$) such that $\lvert C(\mathcal{I}_a) - C(\mathcal{I}_n) \rvert \gg C_{thr}$
\State Train GNN model parameters for $e_1$ epochs.
\State Find a dissimilar problem instance ($n$) such that $\lvert C(\mathcal{I}_a) - C(\mathcal{I}_n) \rvert \gt C_{thr}$
\State Continue training GNN model parameters for $e_2$ epochs.
\end{algorithmic}
\hspace*{\algorithmicindent} Using Loss: $L_{triplet} = [d_{ap} - d_{an} + \alpha]_{+}$
\end{algorithm}

\subsubsection{Feature Extraction}

The MILP formulation represented in Section \ref{sec:preliminaries} does not restrict the order of the decision variables in the objective, nor the number and order of the constraints.
Therefore, a feature extractor needs to be invariant to their order to handle instances of varying sizes.
In Step 5.1.2 of Figure \ref{fig:milptune-overview}, we represent a MILP instance using the bi-partite graph representation from \citep{gasse2019exact}.
Each decision variable is represented as a node, and each constraint is also represented as a node. 
An undirected edge between a decision variable, $v_i$, and a constraint, $a_j$, exists if $v_i$ appears in $a_j$, that is if $\mathbf{A}_{ij} \ne 0$.
Variable nodes have features represented as the variable type (binary, integer or continuous) in addition to its lower and upper bounds.
They are represented as $X \in \mathbb{R}^{n \times d}$, where $n$ is the number of nodes and $d$ is the features dimension.
Constraint nodes have features represented in their (in)equality symbol ($<$, $>$, $=$).
They are represented as $X' \in \mathbb{R}^{m \times a}$, where $m$ is the number of constraints and $a$ is the features dimension.
Edge features represent the coefficients of a decision variable appearing in a constraint, $E \in \mathbb{R}^{n \times m \times e}$, where $e$ is the number of edges.
These features are extracted once before the solver starts the branch-and-bound procedure, namely at the root node.
Therefore, each problem instance has a single graph structure representation before any cuts happen at the root node (part of the heuristics-based algorithms).
While the original representation in \citep{gasse2019exact} has additional features, we only extract the features of the problem instance, and not the solver's state.

\subsubsection{Instance Embedding}

In Step 5.1.3 of Figure \ref{fig:milptune-overview}, we parameterize our distance metric model using a GCN model \citep{kipf2016semi}.
The network structure has four convolutional layers, and the convolutional operator is implemented as defined in \citep{morris2019weisfeiler}.
The network parameters, $\theta_1$ and $\theta_2$, are updated within the end-to-end training procedure where features of the decision variables are updated as:
$x_n \leftarrow \theta_1 x_n + \theta_2 \sum_{m \in \mathcal{N}(n)} e_{n,m} \cdot x'_m$.
Similarly, the features of the constraints are updated as $x'_m \leftarrow \theta_1 x'_m + \theta_2 \sum_{n \in \mathcal{N}(m)} e_{m,n} \cdot x_n$.
Graph embeddings are then passed through batch normalization, max-pooling and attention pooling layers to produce a latent vector which is used for the downstream metric learning loss.

\subsubsection{Model Training}

In Step 5.1.4 of Figure \ref{fig:milptune-overview}, the model is trained end-to-end using the loss function defined in Equation \ref{eq:loss-function}.
The distance function used is the Euclidean distance on the learned metric space.
Remember that the Euclidean distance tends to underperform when calculated on high-dimensional data. However, the non-linear step introduced by the graph neural network enables it to capture relationships between the features of the problem instances that are consistent with their correlation to the final solution costs. In essence, the projection of the problem instance into a learned space allows the Euclidean distance metric to overcome biased outcomes.

The training proceeds for a number of predefined epochs, while ensuring that the loss does not fall to zero by adopting the proposed triplet sampling schedule in Algorithm \ref{algo1}.
The larger the value of $\alpha$, the further positive instances are pushed away from negative ones.
However, choosing a large value of $\alpha$ will make the model set the value of the distance function $d$ as zero.
Thus, $\alpha$ should be tuned for the training procedure.

\subsection{Predicting Configuration Parameters}
\label{sec:sub:predicting-configuration}

In the inference phase (5.2) of Figure \ref{fig:milptune-overview}, the solver is invoked to solve a MILP instance using a given configuration (or default if none is provided).
The goal of this phase is to allow a real-world solver deployment to continue to autonomously improve over time as more configuration parameters are explored.
Thus, we propose a closed-loop system where solutions from real-world problems are continuously saved for future evaluation and use.

\subsubsection{Embedding New Instances}
As motivated earlier in Section \ref{sec:motivation}, we find effective configurations by using a configuration from similar instances in the learned metric space.
Therefore, the first step is to embed (i.e., encode) the new problem instance using the learned model ($M)$ from the training phase.
The embedding time is negligible compared to the solving time as it takes a few milliseconds to extract MILP features and run them through the small GCN.
One advantage of adopting a deep embedding method in our approach is that it is inductive \cite{hamilton2017inductive}, and can generate embeddings for MILP instances of different sizes (i.e., number of decision variables or constraints).
In other words, it does not require re-training the model to accommodate new instances seen in a real deployment.

\begin{algorithm}[t!]
\caption{Predicting Solver Configuration}\label{algo2}
\textbf{Input:} Unseen MILP instance ($\mathcal{I}$) \\
\textbf{Input:} Trained embedding model ($M$) \\
\textbf{Parameters:} \# of nearest neighbors ($k$), \# of predicted configurations ($n$) \\
\textbf{Output:} Predicted solver configuration
\begin{algorithmic}[1]
\State Embed instance $\mathcal{I}$ using $M$.
\State Retrieve previously-solved $k$ nearest neighbors.
\State Select $n$ previously-explored configurations from each retrieved neighbor in a non-descending order according to their associated costs.
\State Return the configuration with the lowest cost.
\end{algorithmic}
\end{algorithm}

\subsubsection{Nearest Neighbor Instances}

A trained model is a model capable of measuring a distance metric between MILP instances.
The final embeddings of the instances are saved in a central store to be used in the prediction step.
Algorithm \ref{algo2} gives the steps performed for predicting a parameters configuration for a new unseen MILP instance. 
In Step 1, the problem instance is first embedded using the trained model.
In Step 2, we perform a nearest neighbor search on the learned metric space.
We introduce two tuning parameters for the prediction: (1) $k$, representing the number of nearest neighbors we want to fetch, and (2) $n$, representing the number of configurations for each neighbor, sorted in a non-descending order by their solutions' costs. 
In Step 3, we retrieve $n$ previously-explored configurations for each of the $k$ neighbors. 
Then, in Step 4, we predict a parameters configuration as the one with the minimum cost.
If $k=1$ and $n=1$, then the algorithm predicts the lowest cost configuration parameters of the nearest neighbor.
In multi-core environments (e.g., cloud), a practitioner may choose to run the solver in parallel using different configuration parameters and gather an ensemble of solutions for the new problem instance.
In this case, $k$ and $n$ can be exposed as hyperparameters for the prediction model.

\subsubsection{Configuration Space Exploration}
As mentioned in Section \ref{sec:related-work}, during the process of identifying similar problem instances, selecting an appropriate parameter configuration remains a challenge.
Essentially, when adequate time is available for exploration, navigate a vast search space requires an exploration strategy for configuration parameters that are most likely to yield good results. 
Given the huge number of potential solver configurations, we term this issue as the exploration problem.
In our approach, we provide initial configurations to the problem instances used for similarity lookup by independently searching the configuration space of each instance with SMAC \cite{lindauer2022smac3}.
The primary goal of SMAC is to find an optimal set of configuration parameters for a given algorithm to minimize a specific performance metric (e.g., MILP objective cost in our context) within a user-defined search space of possible configurations.
SMAC is based on a Bayesian optimization framework that utilizes surrogate models, such as Gaussian Process Regression or Random Forests, to model the objective function.
It employs an acquisition function, such as Expected Improvement (EI), to balance exploration and exploitation during the search process.
SMAC iteratively refines its surrogate model by querying new points in the configuration space space and updating the model with their corresponding objective function values.
This step is performed offline, separate from the training and inference loops.

However, once the model is deployed in a real-world setting, we enable it to evolve by incorporating a feedback loop in which a solver saves its results to the data store.
Each data point consists of a problem instance's embedding, the configuration employed for solving, and the cost obtained from the solver.
Future lookups using KNN can immediately benefit from the newly added data point without retraining the model since similarity is based on the already-learned embeddings.
This design choice allows our method to be deployed in real-world environments without requiring frequent model retraining.
For the implementation details of the data store, refer to Appendix \ref{appendix:data-management}.

%% file: 5_expr.tex
\subsection{Dataset}
\label{sec:sub-dataset}
We used the publicly available dataset from the ML4CO competition \citep{ml4co-competition}. 
The dataset consists of three problem benchmarks.
The first two problem benchmarks (item placement and load balancing) are extracted from applications of large-scale systems at Google, while the third benchmark is extracted from MIRPLIB \--- a library of maritime inventory routing problems\footnote{Link: \url{https://mirplib.scl.gatech.edu/instances}}.
The item placement and load balancing benchmarks contain 10,000 MILP instances for training (9,900) and testing (100), while the anonymous problem contains only 118 instances (98 and 20 for training and testing, respectively).
The dataset is available to download from the ML4CO competition website\footnote{Link: \url{https://github.com/ds4dm/ml4co-competition}} with a full description on the problems formulation and their sources.
Even the smallest of these problems are extremely hard to solve to optimality.
For example, after 48 hours of solving time using SCIP, an instance of the Item Placement dataset was not solved to optimality on a high-end machine (Section \ref{sec:sub-runtime-env}).
In fact, after 2 hours, the solver reports a gap of 22.00\% and a search progress completion of 23.05\%.
After 12 hours, the solver reports a gap of 14.00\% and a search progress completion of 32.05\%.
After 48 hours, the solver reports a gap of 10.28\% and a search progress completion of 35.60\%.
In this section, we show some statistics on the dataset and reflect on how they affect our approach of metric learning.

\begin{table}[t]
    \renewcommand{\tabcolsep}{3pt}
  \caption{Dataset Statistics}
  \label{tab:dataset-stats}
  \centering
  \begin{tabular}{lrrrrrr}
    \toprule
    & \multicolumn{3}{c}{\# Decision Variables} & \multicolumn{3}{c}{\# Constraints}                    \\
    \cmidrule(r){2-7}
    \textbf{Benchmark} & Count & Avg. & Median & Count & Avg. & Median \\
    \midrule
    \textbf{Item Placement} & 195  & 195 & 195 & 1,083 & 1,083 & 1,083     \\
    \textbf{Load Balancing} & 61,000  & 61,000 & 61,000 & 64,081\---64,504 & 64,307 & 64,308    \\
    \textbf{Anonymous}      & 1,613\---92,261  & 33,998 & 4,399 & 1,080\---12,6621 & 43,373 & 2,599     \\
    \bottomrule
  \end{tabular}
\end{table}

Table \ref{tab:dataset-stats} shows the number of decision variables and constraints in each benchmark.
All instances in the Item Placement benchmark have the same number of decision variables and constraints.
The Load Balancing benchmark has the same number of decision variables, but the number of constraints varies within a small range.
The Anonymous benchmark exhibits a large variance in both the number of decision variables and constraints.
For a MILP solver, a high variance in the number of decision variables or constraints has a direct impact on its solution.
It also affects the learned embeddings of these instances.
While the high variance gives more discriminative power to the model ($M$), it does not directly serve the purpose of finding a configuration for new instances using the nearest neighbor.
The reason is that the nearest neighbor might indeed not be close in distance in the learned metric space, and the predicted parameters configuration would not be directly correlated to the solver's solution.
Therefore, it is critical that the definition of ``same distribution'' instances include the number of decision variables and constraints for the purpose of finding a parameters configuration using metric learning.

\subsection{Experimental Setup}
In this section, we provide details on our runtime environment, the data utilized for training, and the training methodology.
Subsequently, we design a series of experiments to evaluate the effectiveness of our approach, both in terms of learning meaningful MILP embeddings and its influence on the final solution's cost when employing the complete system illustrated in Figure\nobreakspace\ref{fig:milptune-overview}.
First, in Section \ref{sec:results-embedding}, we delve into the learned MILP embeddings and examine their correlation with the final solution costs when solved in the same environment.
Next, in Section \ref{sec:results-accuracy}, we explore the precision of the predicted configurations in identifying suitable configuration parameters.
Finally, in Section \ref{sec:results-baselines}, we compare our method with existing approaches for selecting parameter configurations and discuss the implications of learning improved similarity models as they relate to the predicted costs after solving.

\vspace{0.1in}
\subsubsection{Runtime Environment}
\label{sec:sub-runtime-env}
The experimental results are obtained using a machine with Intel Xeon E5-2680 2x14cores@2.4 GHz, 128GB RAM, and a Tesla P40 GPU.
The model was developed using PyTorch (v1.11.0+cu113)\nobreakspace\citep{paszke2019pytorch}, Pytorch Geometric (v2.0.4)\nobreakspace\citep{Fey-Lenssen-2019}, and PyTorch Metric Learning (v1.3.0) \citep{musgrave2020pytorch}.
We used Ecole (v0.7.3) \citep{prouvost2020ecole} for graph feature extraction, convolution operators modified and adopted from\nobreakspace\citep{velantin21}, PySCIPOpt (v3.5.0) \citep{pyscipopt} as the MILP solver, and SMAC3 (v1.2) \citep{lindauer2022smac3} for the offline configuration space search.

\subsubsection{MILP Triplet Sampling}
Given the training dataset, we run the MILP solver on all instances using the default parameters configuration of the solver with a time limit of 15 minutes as suggested by \cite{ml4co-competition}.
The total number of solved instances by the end of the time limit were 2599, 1727 and 38 for the item placement, load balancing and anonymous benchmarks, respectively.
This represents 26\%, 17\% and 38\% of the training benchmarks, respectively.
We implemented the triplet sampling schedule as discussed in Section \ref{sec:methodology}, where hard negative sampling was used for the first 50 epochs, and the training continues for 100 epochs in total.
We used a batch size of 256 for the item placement, 64 for load balancing, and the full 98 instances for the anonymous benchmark.

\begin{figure}
\renewcommand{\tabcolsep}{0pt}

\begin{tabular}{cccc}
\vspace{-0.2in}
\rotatebox[origin=c]{90}{Before Embedding} &
\includegraphics[clip, width=1.6in, height=1.55in, valign=m]{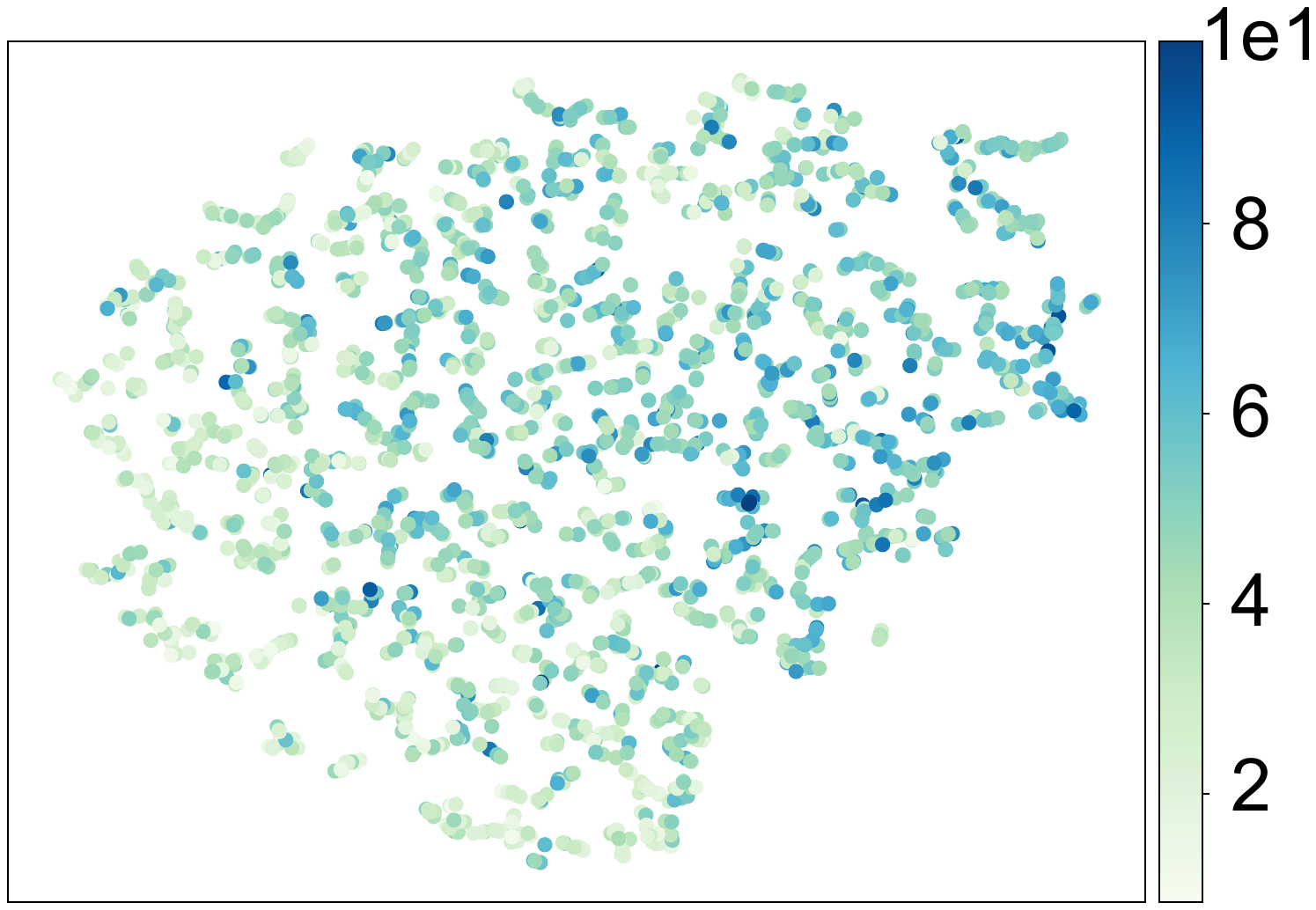} &
\includegraphics[clip, width=1.6in, height=1.55in, valign=m]{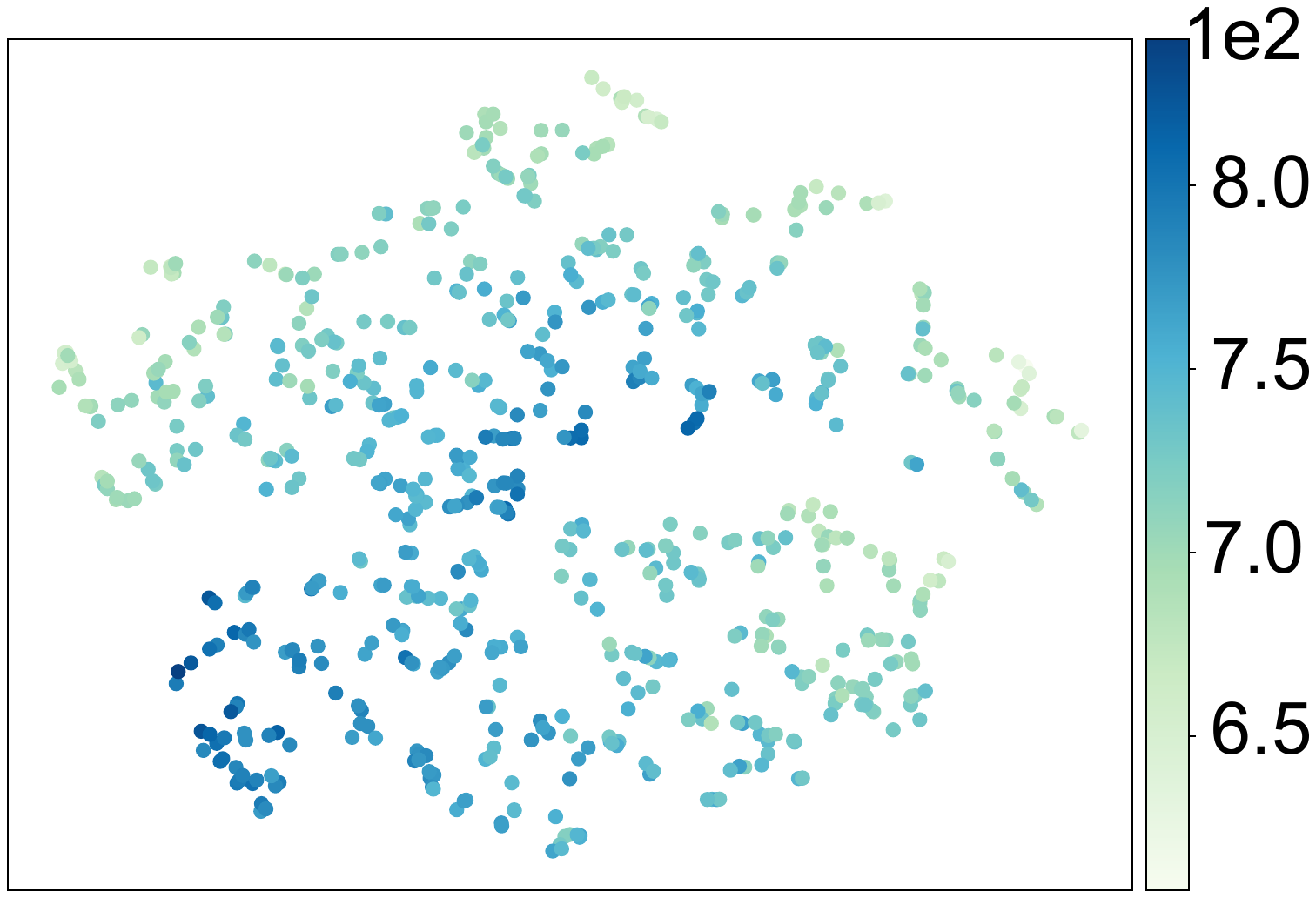} &
\includegraphics[clip, width=1.6in, height=1.55in, valign=m]{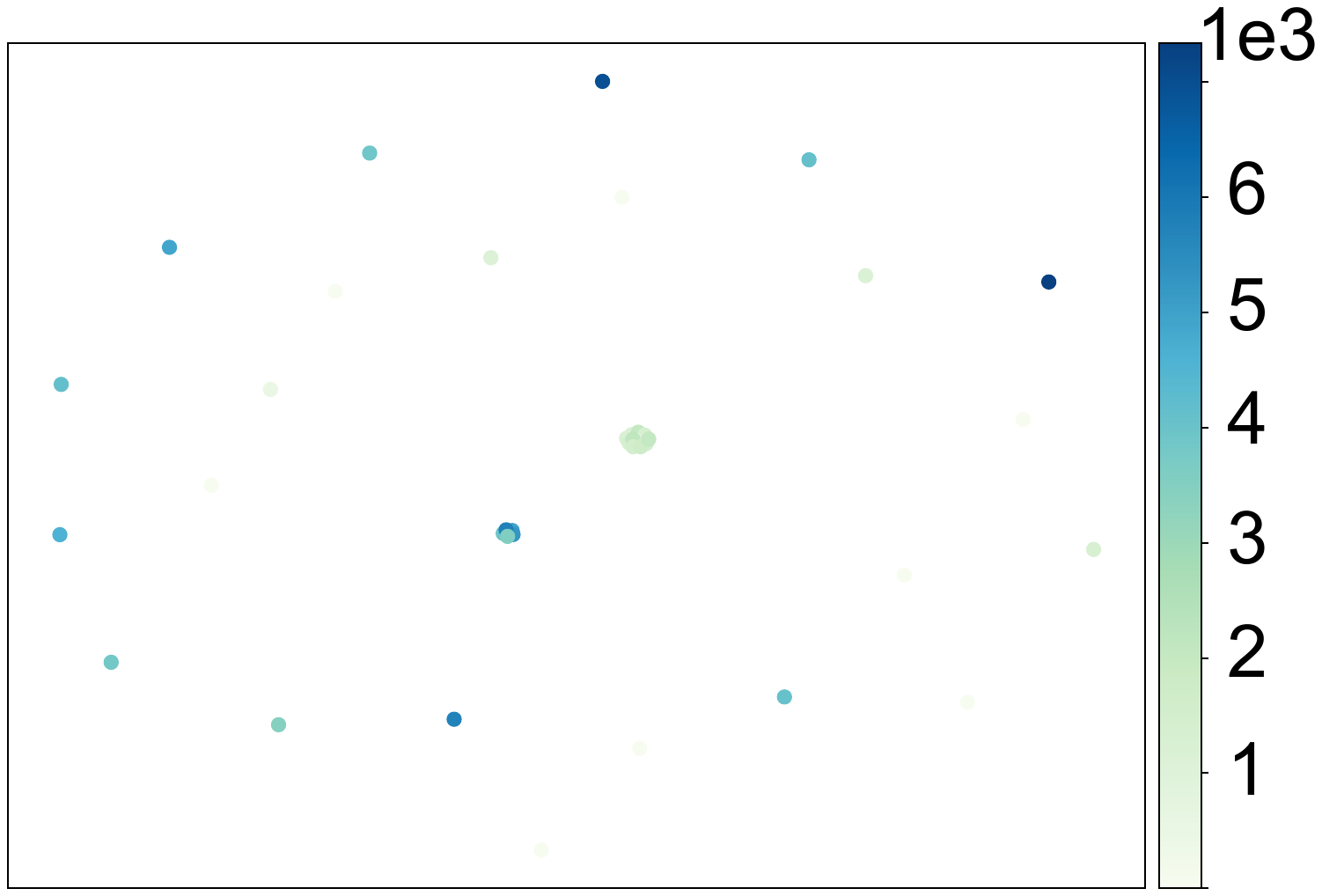}
\\
\vspace{-0.2in}
\rotatebox[origin=c]{90}{Shallow Embedding} &
\includegraphics[clip, width=1.6in, height=1.55in, valign=m]{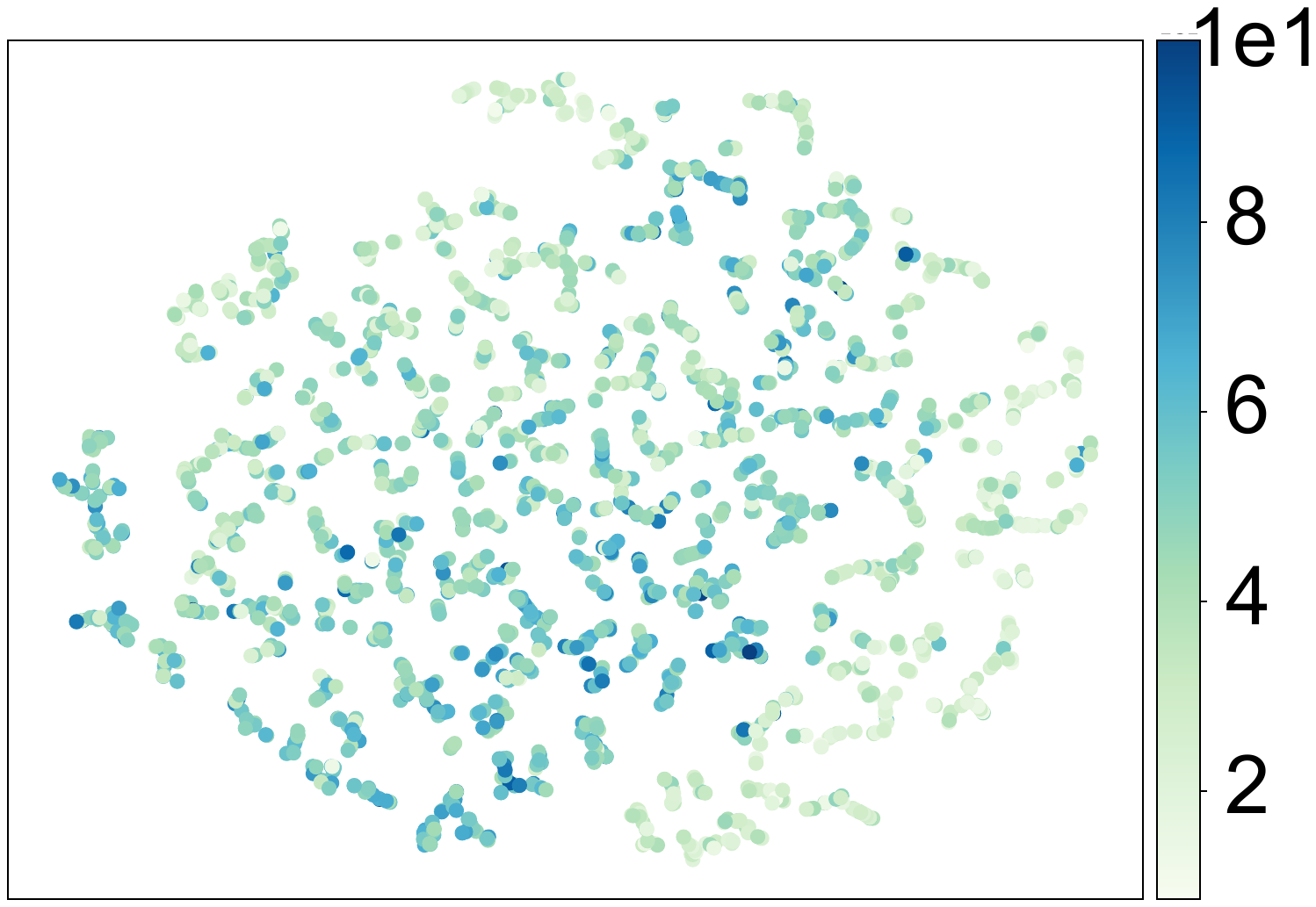} &
\includegraphics[clip, width=1.6in, height=1.55in, valign=m]{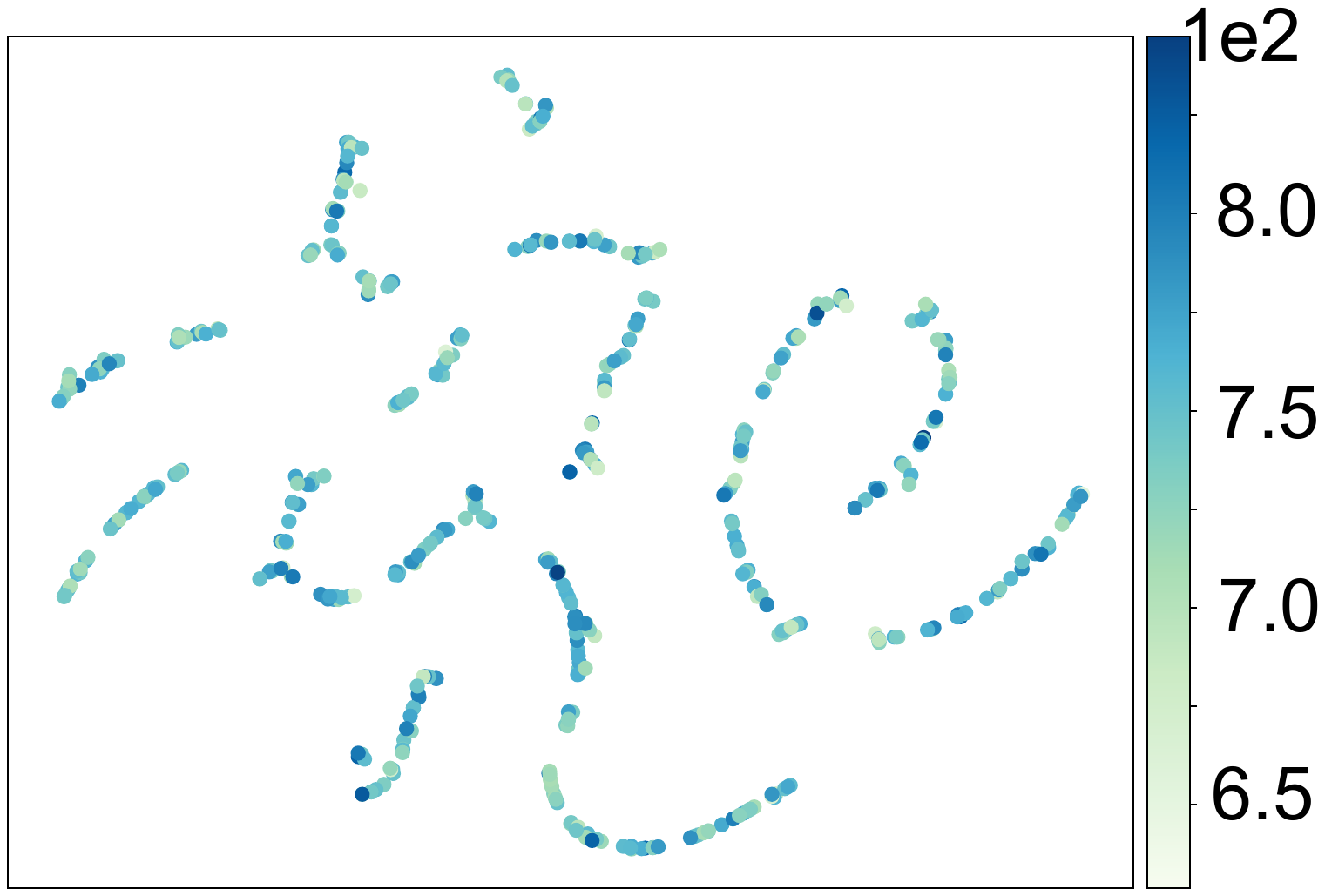} &
\includegraphics[clip, width=1.6in, height=1.55in, valign=m]{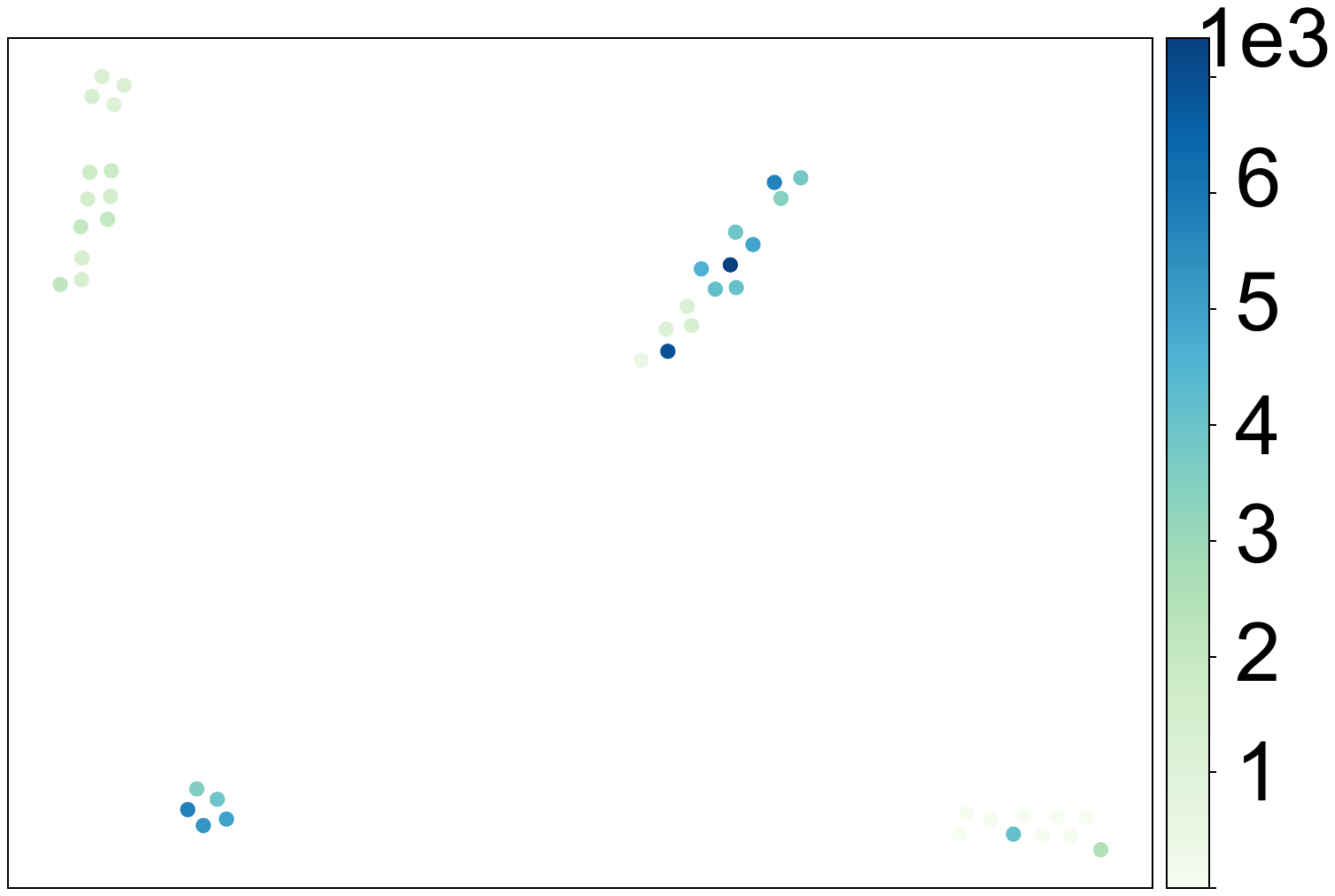}
\\
\vspace{-0.14in}
\rotatebox[origin=c]{90}{Deep Embedding} &
\includegraphics[clip, width=1.6in, height=1.55in, valign=m]{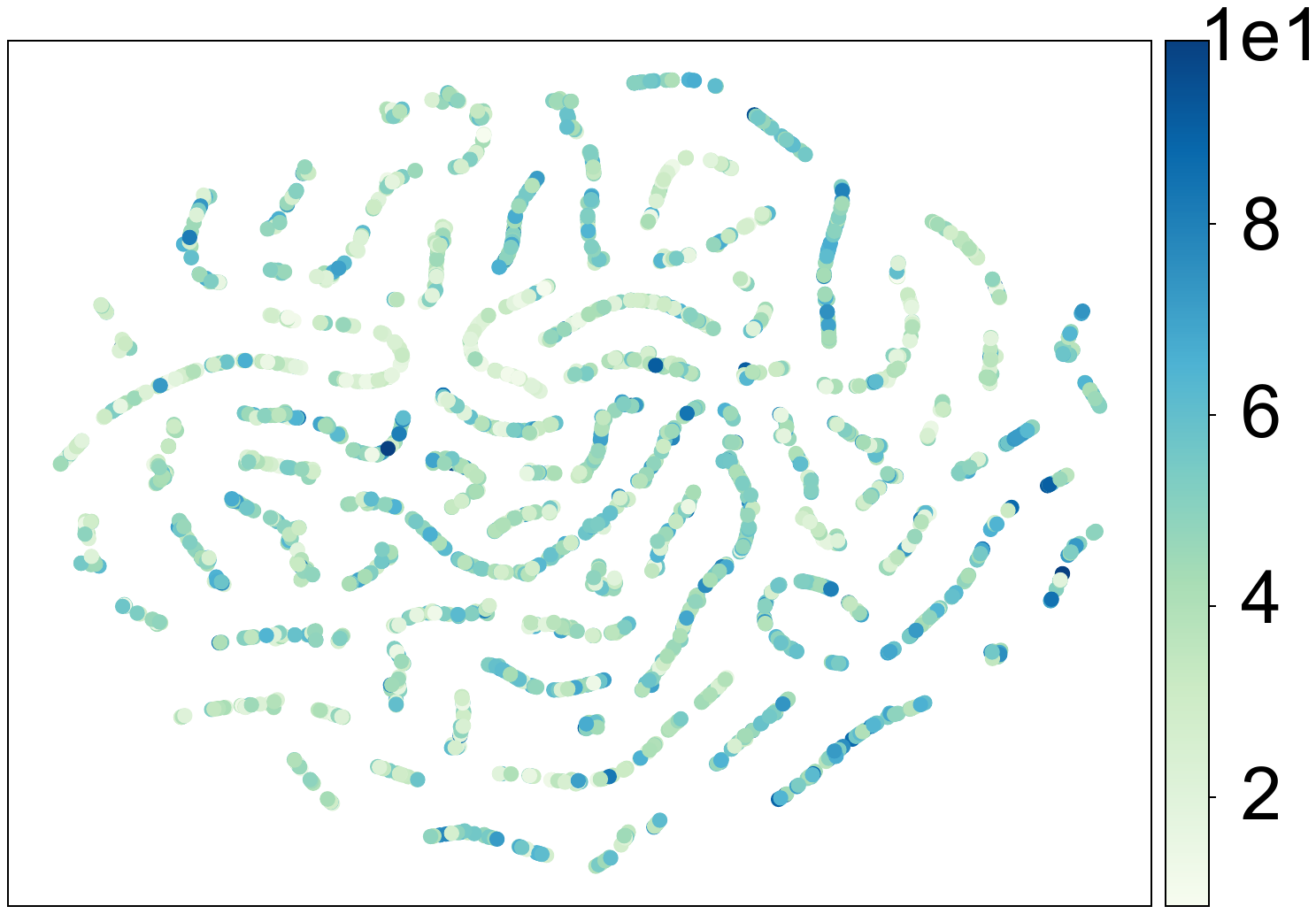} &
\includegraphics[clip, width=1.6in, height=1.55in, valign=m]{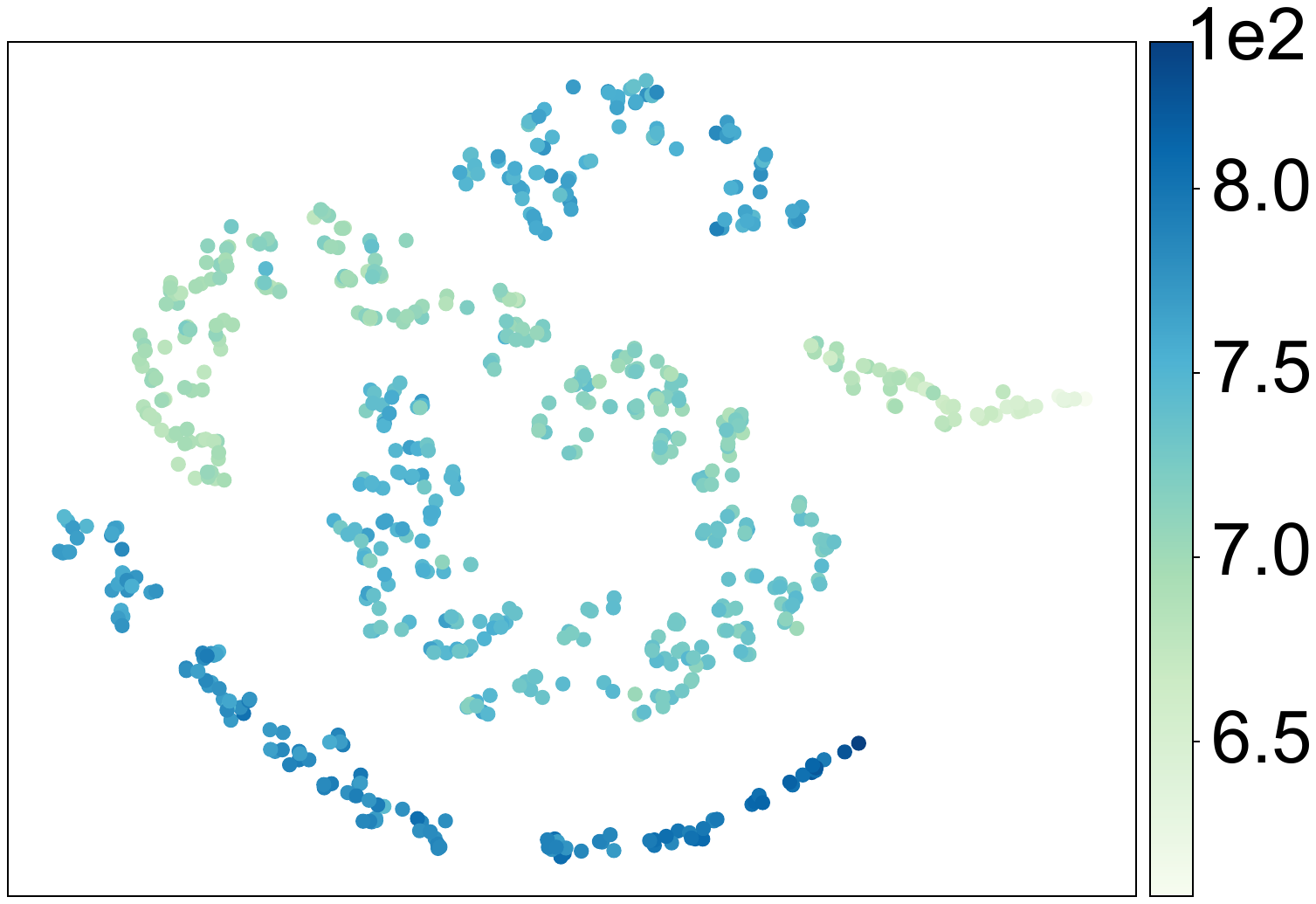} &
\includegraphics[clip, width=1.6in, height=1.55in, valign=m]{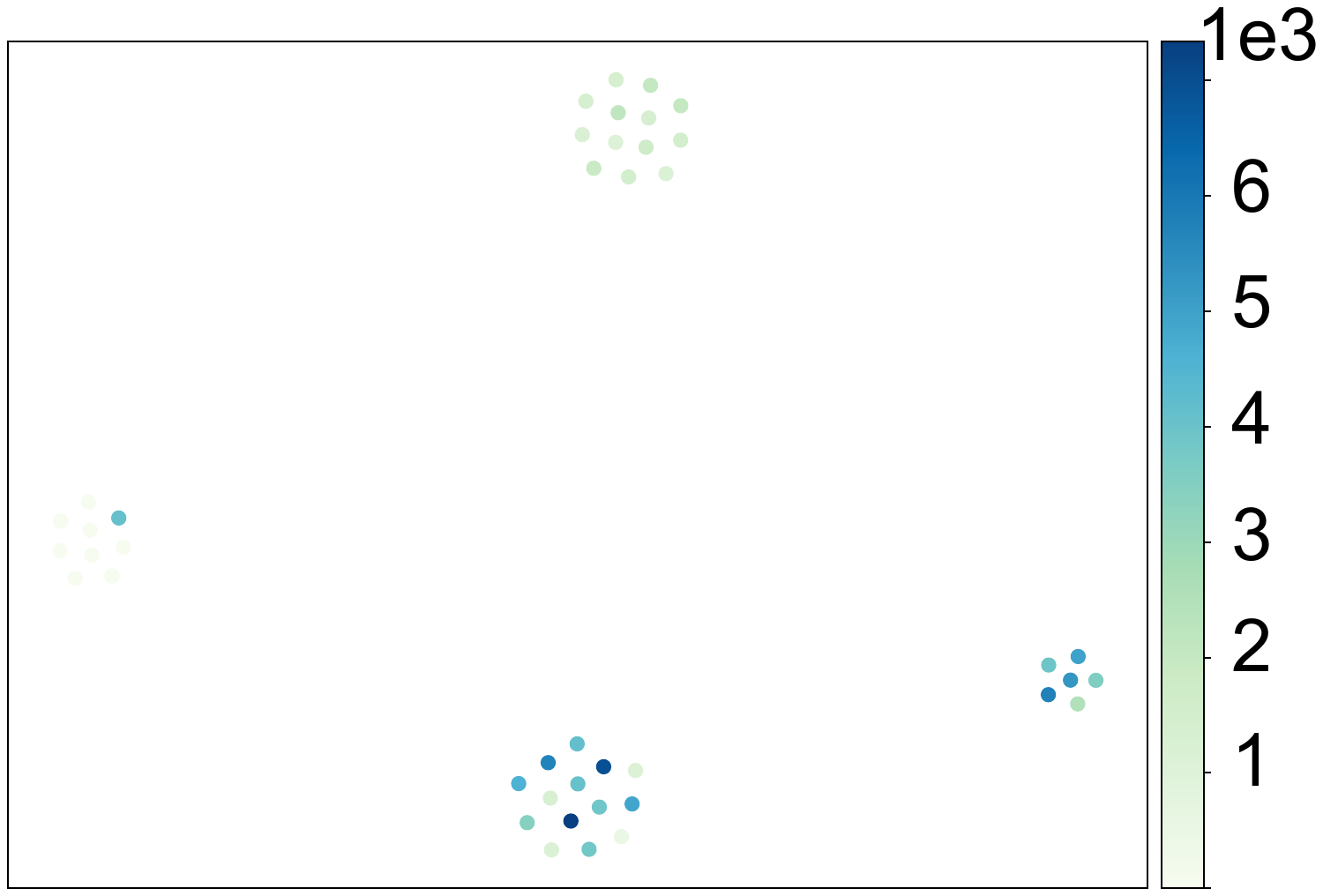}
\\
& Item Placement & Load Balancing & Anonymous
\\
\end{tabular}
\vspace{0.1in}
\caption{Vector representations of MILP problem instances visualized using t-SNE \citep{tsne}. Each point is a problem instance where the color denotes its solution's cost using SCIP's \cite{pyscipopt} default configuration. Item placement, load balancing and anonymous represent benchmarks from the ML4CO dataset \cite{ml4co-competition}. The first row (Before Embedding) represents the feature vector of instances before learning any similarities (i.e., random). The second row (Shallow Embedding) encodes the vector representation using \cite{xu2011hydra}. The third row (Deep Embedding \-- Our Method) encodes the vector representation in the learned embedding space. In the item placement benchmark, shallow embedding does not offer any discriminative capability. In the load balancing benchmark, it could cluster problem instances, but clusters are not correlated with the final solver's costs. Shallow embedding uniquely embeds the anonymous benchmark and the embedding is correlated to the final costs. The discriminative power of deep embedding is evident in the three benchmarks, where similarity is directly correlated with the cost after running the solver using the default configuration.}
\label{fig:network-embedding}
\end{figure}

\subsubsection{Model Training}

The model consists of a graph neural network of four layers with 64 as the dimension of the hidden layers.
It is trained for each benchmark separately in order for the triplet sampling and training to run on data coming from the same distribution.
The output from the convolutional layers is passed into a batch normalization layer, followed by a max pooling layer and an attention pooling layer.
The output embedding size is set to 256.
We set $\alpha = 0.1$ in the loss function.

\subsection{Instance Embedding}
\label{sec:results-embedding}

We visualize the instance embeddings of the GNN before and after model training and compare it to using shallow embeddings in Figure \ref{fig:network-embedding}. 
The color bar represents the cost of the solution using the default configuration parameters.
The shallow embedding vector encodes presolving statistics as in Hydra-MIP\nobreakspace\citep{xu2011hydra}, which include the problem size, the minimum, maximum, average and standard deviation of the objective coefficients ($\mathbf{c}$) and the constraints coefficients ($\mathbf{A}$, $\mathbf{b}$).
While Hydra-MIP's shallow embedding includes more features such as the cutting planes usage and the branch-and-bound tree information, such information is not available before running the solver\footnote{The implementation of shallow embedding is provided in the supplementary material. There is no publicly available implementation of Hydra-MIP.}.
From Figure \ref{fig:network-embedding}, we observe that in the item placement benchmark, shallow embeddings do not offer any discriminative power to the problem instances.
In the load balancing benchmark, shallow embeddings could indeed cluster problem instances, but clusters are not correlated with the final solver's costs.
In the anonymous benchmark, instances with similar costs were clustered close to each other, which gives shallow embedding a discriminative power in this case.
Analyzing this result in light of the dataset statistics (Table \ref{tab:dataset-stats}), we see that the anonymous benchmark has a high variance in the number of decision variables and constraints.
Therefore, a feature vector that includes aggregated values could distinguish the problem instances.
On the other hand, where item placement has the same number of decision variables and constraints, a shallow feature vector could not capture the graph connectivity properties, nor the coefficients values.
Between these two cases, the load balancing benchmark has the same number of decision variables, while the number of constraints do not have a high variance (64,081 to 64,504 constraints).
Shallow embedding was able to cluster problem instances, but its clusters were not correlated to the final solver's costs.
The learned embeddings in our method is discriminative in the three benchmarks.

\begin{figure}[t]
    \begin{subfigure}{0.33\textwidth}
        \includegraphics[clip, scale=0.25]{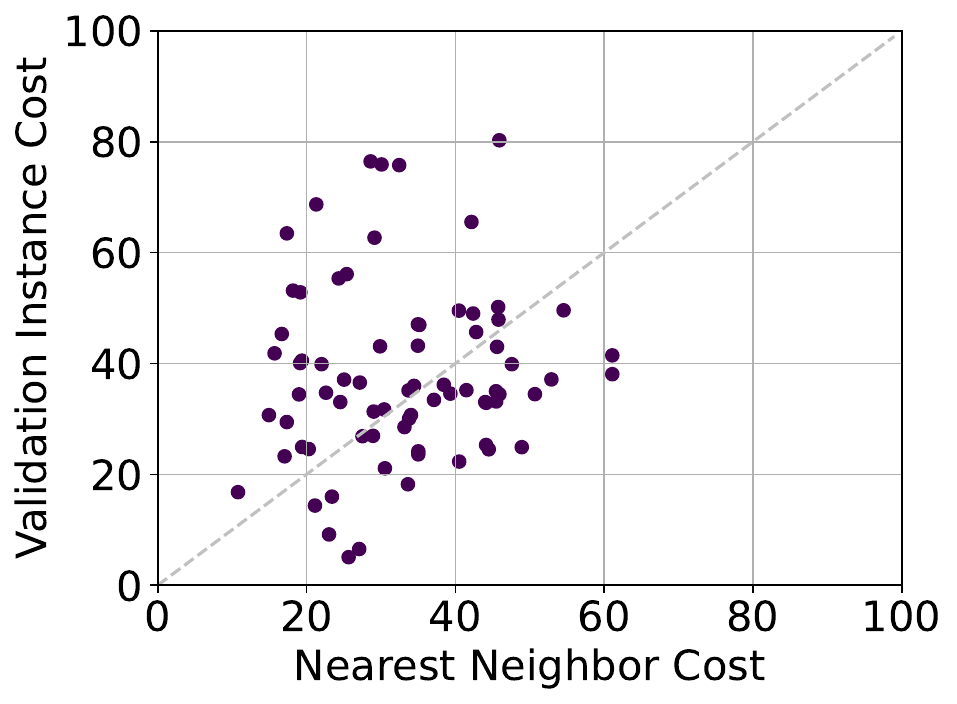}
        \caption{\centering Item Placement  MAE=18.07}
    \end{subfigure}
    \begin{subfigure}{0.33\textwidth}
        \includegraphics[clip, scale=0.25]{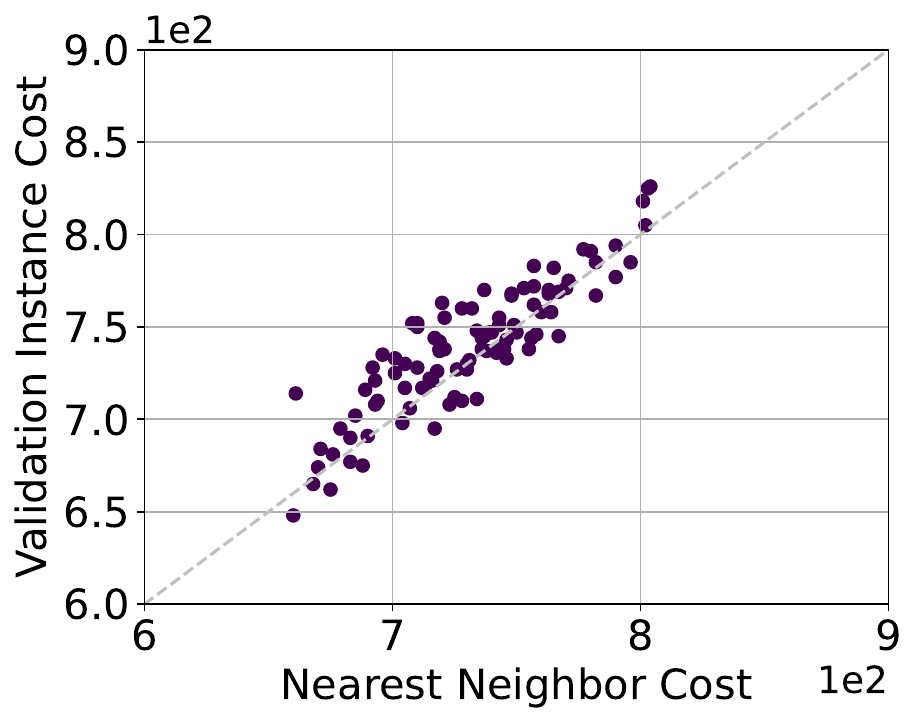}
        \caption{\centering Load Balancing  MAE=14.46}
    \end{subfigure}%
    \begin{subfigure}{0.33\textwidth}
        \includegraphics[clip, scale=0.25]{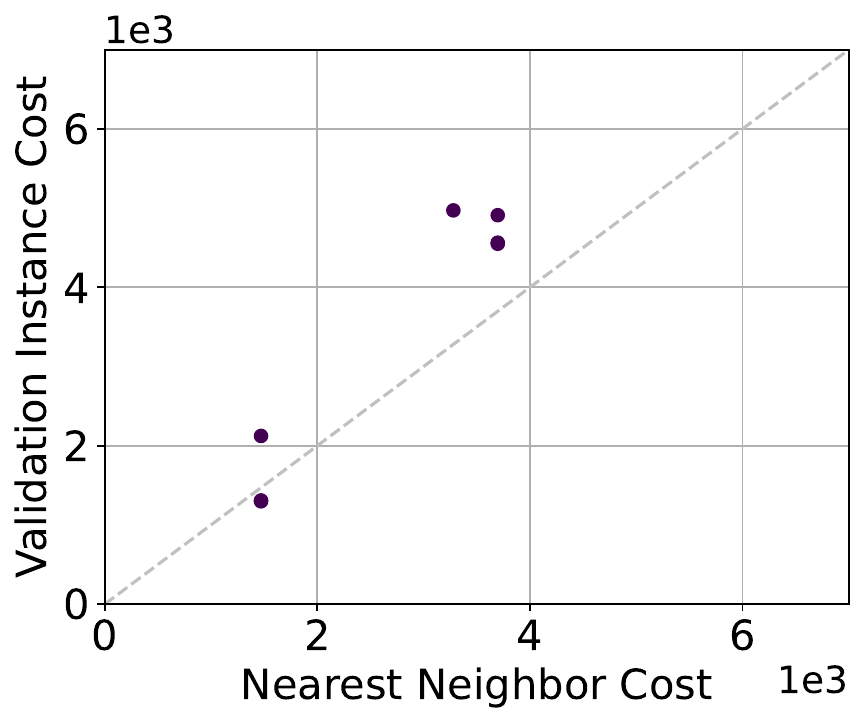}
        \caption{\centering Anonymous  MAE=801.36}
    \end{subfigure}
  \caption{Cost (primal bound) of using the predicted configuration from the nearest neighbor in the learned metric space (x-axis) as compared to the actual cost after using it for the validation instance (y-axis).}
  \label{fig:prediction-accuracy}
\end{figure}

\subsection{Prediction Accuracy}
\label{sec:results-accuracy}

A key question in our approach is whether the nearest neighbor in the embedding space would exhibit a similar solver behavior when using its parameter configuration.
Here, we embed the validation instances using our trained model, and then obtain a parameters configuration from the nearest neighbor.
Then, we run the solver using the predicted parameters configuration on the validation instances (T=15mins).
Figure \ref{fig:prediction-accuracy} plots the solution's cost of the predicted parameters configuration from the nearest neighbor (x-axis) vs. its solution's cost on the validation instance (y-axis). 
It shows that there is indeed a correlation between the final cost of the solution using the predicted parameters configuration, and the stored nearest neighbor cost using that configuration.
The mean absolute errors (MAE) were 18.07, 14.46, and 801.36 for item placement, load balancing and anonymous, respectively.
This correlation proves that, in reality, similar MILP instances based on the learned metric space expose similar solver behaviors yielding similar solution costs.
In other words, finding a good parameters configuration for one problem instance can be used for similar instances without repeating an exhaustive search at deployment time.

\subsection{Comparing to Baselines}
\label{sec:results-baselines}

We compare our method against existing approaches in Table \ref{table:detailed-results}.
The first baseline is using SCIP's default configuration, which is usually used by most practitioners.
In addition, we obtain an incumbent configuration by performing a configuration space search on the training instances using SMAC \citep{lindauer2022smac3}.
We perform this search for each benchmark separately.
Although the number of unique configurations explored was 51012 over a period of over 12000 core-hours, this represents a small subset of the configuration space.
Moreover, we implement Hydra-MIP \citep{xu2011hydra}, which uses a statistics-based vector for instance embedding and pair-wise weighted random forests for configuration selection.
In Hydra-MIP, the pairwise weighted random forests (RFs) method is used to select amongst $m$ algorithms for solving the instance, by building $m \cdot (m-1)/2$ RFs and taking a weighted vote.
In our processed dataset, the number of \textit{unique} configurations explored offline using SMAC are 22580, 27971 and 461 for the item placement, load balancing and anonymous training benchmarks, respectively.
Among those, the number of \textit{unique} configurations that worked \textit{best} on their respective instances (excluding unsolved instances) are 4325, 3987 and 53.
As a result for the Hydra-MIP approach, building the portfolio by performing algorithm selection using pairwise RFs is computationally infeasible (memory and compute).
For example, in the item placement benchmark, a total of $4325 \times 4324 / 2 = 9350650$ RFs are needed.
To obtain results for Hydra-MIP, we selected a subset of the top 100 performing configurations in the item placement and the load balancing benchmark, and used all 53 best configurations of the anonymous benchmark.
Lastly, we compare against using the shallow embedding from Hydra-MIP with KNN, which avoids the scalability limitation of RFs.
Table \ref{table:detailed-results} reports the number of instances solved with the lowest cost in each method, along with the average cost improvement over using the default configuration. 
We see that our method predicts configurations that solve more instances, with up to 38\% improvement in the cost of the objective function (confidence level of 95\%).

\begin{figure}
    \centering
    \begin{subfigure}{0.33\textwidth}
        \includegraphics[clip, width=1.65in, height=1.2in]{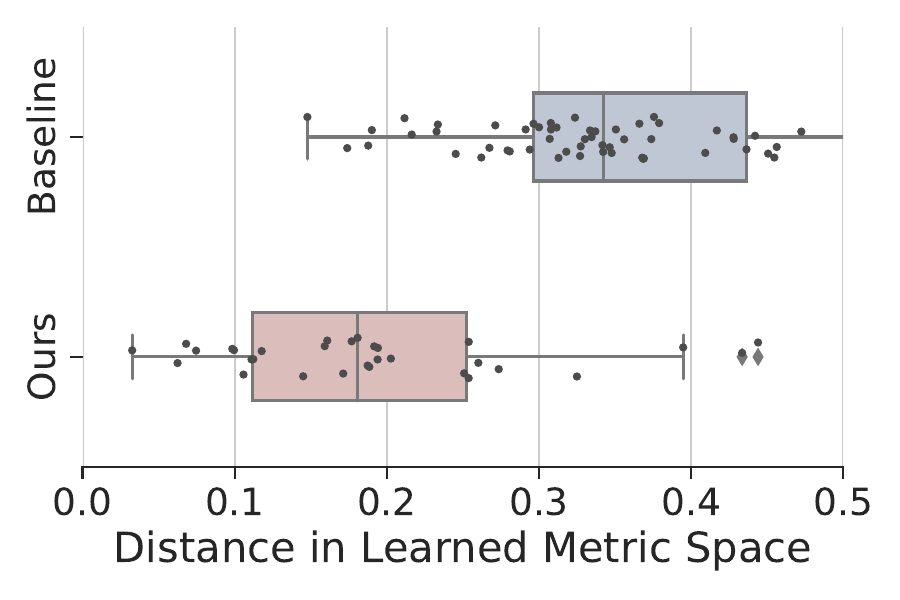}
        \centering
        \caption{Item Placement}
    \end{subfigure}
    \begin{subfigure}{0.33\textwidth}
        \includegraphics[clip, width=1.65in, height=1.2in]{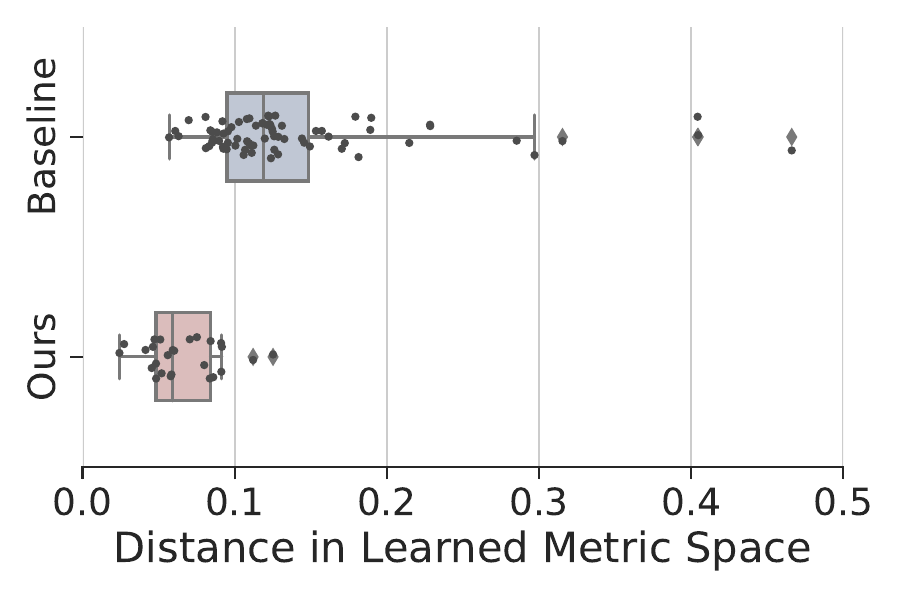}
        \centering
        \caption{Load Balancing}
    \end{subfigure}%
    \begin{subfigure}{0.33\textwidth}
        \includegraphics[clip, width=1.65in, height=1.2in]{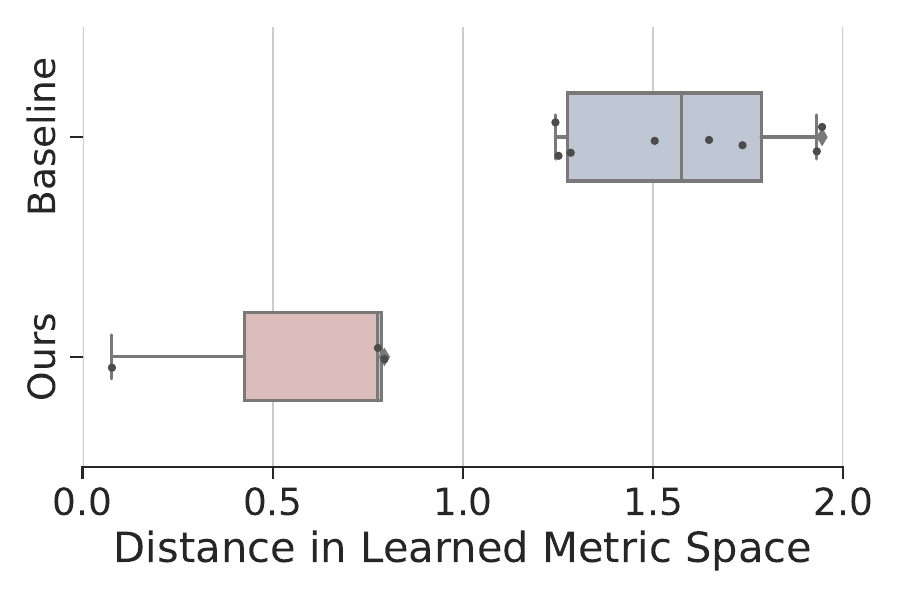}
        \centering
        \caption{Anonymous}
    \end{subfigure}
  \caption{Similarity in the learned embedding space. The x-axis represents the distance between the validation instance and its nearest neighbor in the learned metric space. The y-axis represents the method that gives a better solution. The closer the nearest neighbor to the validation instance, the better the predicted configuration by our method. Baseline represents the method with the lowest cost amongst the default configuration, SMAC's incumbent, Hydra-MIP and shallow embedding.}
  \label{fig:distance-switch}
\end{figure}

Moreover, we investigate how our method brings instances with similar final costs close to each other by plotting the winning predictions against their distance from their neighbors in the learned embedding space.
In Figure \ref{fig:distance-switch}, the x-axis represents the distance between the validation instance and its nearest neighbor, while the y-axis represents the method that offers a better parameters configuration. 
We observe that the smaller the distance between the validation instance and its nearest neighbor in the learned embedding space, the more probable the neighbor's parameters configuration to yield a better solution than other baselines.
In other words, our method correlates the similarity of the learned embedding to the final solution costs.

\begin{table}[t]
\renewcommand{\tabcolsep}{3pt}
      \caption{Our Method vs. Existing Approaches. In the dataset \citep{ml4co-competition}, there are 100, 100 and 20 test instances for the item placement, load balancing and anonymous benchmarks, respectively. Imprv. represents the average solution's cost improvement over the cost obtained using the default configuration of the SCIP solver. Cost is the value of the MILP objective function using the solution found by the solver. Since the smallest problem instance takes several days to solve to optimality, we limit the runtime to 15mins as suggested in\nobreakspace\citep{ml4co-competition}\nobreakspace(Section\nobreakspace\ref{sec:sub-dataset}). Wins represents the number of instances that a method solved with the lowest cost within the time limit. Shallow embedding + KNN (our baseline) uses the same embedding vector as \citep{xu2011hydra}. Deep embedding (our method) is evaluated at $k=1$ and $n=1$ (See. Algo. \ref{algo2}).}
  \label{table:detailed-results}
  \centering
  \begin{tabular}{lrrrrrr}
    \toprule
    & \multicolumn{2}{c}{Item Placement} & \multicolumn{2}{c}{Load Balancing} & \multicolumn{2}{c}{Anonymous}                   \\
    \cmidrule(r){2-7}
    Configuration & Wins & Imprv. $\downarrow$ & Wins & Imprv. $\downarrow$  & Wins & Imprv. $\downarrow$ \\
    \midrule
    No Solution Found & 0 & - & 0 & - & 11 & - \\
    Default SCIP Config & 1 & - & 34 & - & 1 & - \\
    Incumbent from SMAC \citep{lindauer2022smac3} & 8 & 0.24$\pm$0.16 & 4 & 0.01$\pm$0.03 & 1 & 0.01$\pm$0.00 \\
    Hydra-MIP \citep{xu2011hydra} & 10 & 0.25$\pm$0.09 & 17 & 0.02$\pm$0.01 & 0 & - \\
    Shallow Embedding + KNN & 16 & 0.17$\pm$0.08 & 5 & 0.04$\pm$0.06 & 3 & 0.11$\pm$0.02 \\
    \textbf{Deep Embedding + KNN} & \textbf{65} & \textbf{0.38$\pm$0.06} & \textbf{40} & \textbf{0.04$\pm$0.03} & \textbf{4} & \textbf{0.26$\pm$0.07} \\
    \bottomrule
  \end{tabular}
\end{table}

%% file: 6_discussion.tex
\textbf{Generalizing to Other Solvers.}
MILP solvers expose different configuration parameters for their internal algorithms.
For example, while SCIP exposes over 2500 parameters\footnote{\url{ https://www.scipopt.org/doc/html/PARAMETERS.php}}, CPLEX exposes 182 parameters\footnote{\url{https://www.ibm.com/docs/en/icos/12.8.0.0?topic=cplex-list-parameters}} and Gurobi exposes ~100 parameters\footnote{\url{https://www.gurobi.com/documentation/9.0/refman/parameters.html}}.
Due to the different algorithm implementations, only a small subset of parameters have an exact match across all solvers.
SCIP has been used in this work for a number of reasons: (1) it is a stable open-source solver and its algorithms are comprehensively documented, while commercial tools hide their implementation details (2) it exposes a large number of configuration parameters to tune, and (3) it has been used in previous related works \citep{gasse2019exact, kruber2017learning, prouvost2020ecole, velantin21}.

In order to generalize our method to other solvers, it is important to note that a solution’s cost depends primarily on: (1) the problem instance, (2) the solver used (including the specific solver version), (3) the time limit, (4) the hardware resources given to the solver (cores and memory), in addition to (5) the configuration parameters. 
For the purpose of learning similarity between MILP instances, the solver’s costs are used as a subjective measure of the similarity between two instances that use the same solver version, time limit, hardware resource, and configuration parameters.
Replacing the solver with another solver is possible for the sake of getting costs that could be used to measure the similarity between different MILP instances.
However, it is critical to fix all parameters of the solving environment (hardware, solver tool and its version, time limit, configuration parameters) in order for the cost to be representative of the similarity.
Once a similarity measurement is established, two similar instances in one solver’s environment could potentially be used to determine that these two instances will have similar costs in another solver’s environment.
However, we have not investigated this path in the scope of this study and will leave it for future work.

\noindent\textbf{Limitations.}
Our adoption of metric learning in configuring MILP solvers relies on data that represent the same problem being solved repeatedly.
In Section \ref{sec:sub:triplet-sampling}, we sampled triplets of anchor-positive and anchor-negative based on $C_{thr}$.
It is infeasible to identify similar triplets if problem instances are coming from different distributions where the range of their costs varies significantly.
For example, the cost range of the Item Placement benchmark is [0, 100] while the cost range of the Load Balancing is [500, 1000].
While finding a dissimilar pair is straightforward (e.g., one instance from each benchmark), it is hard to find a similar pair where the cost difference is $\lt C_{thr}$.
This means that in order to train a deep embedding model for learning MILP similarity, the MILP formulation needs to represent a problem being solved repeatedly, which is materialized in the number of decision variables or constraints in the problem, as well as the range of their solutions' costs.

\noindent\textbf{Reproducibility.}
In Section \ref{sec:expr}, we refer the reader to the original dataset to download.
A link to the processed dataset (learned embeddings) is available in the supplementary material.
In addition, we describe our setup for training and the pipeline architecture.
The source code along with the training implementation is available in the supplementary material.

%% file: 7_conclusion.tex
In this study, we tackle the challenge of selecting configuration parameters for Mixed-Integer Linear Programming (MILP) solvers.
We propose an instance-aware method that predicts parameter configurations for new problem instances using deep metric learning.
Our approach aims to \textit{learn} a reliable similarity metric between MILP instances, which correlates with the solver's behavior, specifically the final solution's cost when using the same parameter configuration.
Our method offers several advantages compared to existing approaches.
Firstly, it provides more discriminative power to instance features by leveraging the capabilities of deep metric learning, which allows for a better understanding of the relationships between problem instances and their optimal parameter configurations.
Secondly, our method predicts parameter configurations that lead to improved solutions, with up to a 38\% enhancement in performance. 
This improvement highlights the effectiveness of using a learned similarity metric in guiding the selection of solver configurations.
Finally, the system we have designed enables our method to be deployed in real-world environments and continue to evolve through offline exploration without the need for frequent retraining of the learned models.
This adaptability enhances the practical applicability of our approach, as it can seamlessly adapt to new problem instances and improve its performance over time without significant additional computational overhead.

In the future, we plan to investigate the potential of utilizing the learned similarity metric to generate new parameter configurations that were not encountered during the offline search.
By doing so, we aim to explore novel configuration space search algorithms based on the learned similarity of problem instances.
This approach could lead to more efficient and effective methods for optimizing solver performance across a wide range of problem instances, further enhancing the overall effectiveness of MILP solvers in diverse real-world applications.

%% file: appendix_1.tex
In order to offer a seamless integration of our method in existing environments, a data store is required to save the results from the offline configuration space search.
In this work, we use MongoDB\footnote{Link: \url{https://www.mongodb.com/}} for that purpose.
For each benchmark, we create a collection that contains records for each problem instance in that dataset. Listing \ref{codesample} shows the schema used for each instance.
It keeps track of configurations explored for that instance along with their costs.
In addition, it records the embedding vector of the instance in order to be searched later with the nearest neighbor algorithm.
The parameters presented in the listing are the ones that were used for the configuration space exploration using SMAC \cite{lindauer2022smac3}.
A detailed description of the definition of these parameters can be found in their official documentation\footnote{Link: \url{https://www.scipopt.org/doc/html/PARAMETERS.php}}.
As discussed in Section \ref{sec:methodology}, the metric learning approach does not limit the number of configuration parameters explored offline.
It also does not limit which parameters are explored since it focuses on learning an embedding space where similarity between instances can be quantified reliably.
Thus, it is possible to learn a model for similarity once and keep expanding the offline configuration space search without requiring to re-train the model.

\begin{lstlisting}[language=Python, caption=Problem Instance Record, label=codesample]
instance_record = {
    "configs": [
        {
            "seed": 0,
            "cost": 0,
            "time": 0,
            "params": {
                "branching/scorefunc": "s",
                "branching/scorefac": 0.167,
                "branching/preferbinary": False,
                "branching/clamp": 0.2,
                "branching/midpull": 0.75,
                "branching/midpullreldomtrig": 0.5,
                "branching/lpgainnormalize": "s",
                "lp/pricing": "l",
                "lp/colagelimit": 10,
                "lp/rowagelimit": 10,
                "nodeselection/childsel": "h",
                "separating/minortho": 0.9,
                "separating/minorthoroot": 0.9,
                "separating/maxcuts": 100,
                "separating/maxcutsroot": 2000,
                "separating/cutagelimit": 80,
                "separating/poolfreq": 10
            }
        },
        :   # all configurations explored offline
    ],
    "bipartite": {
        "vars_features": [...],
        "cons_features": [...],
        "edge_features": [...]
    },
    "embedding": [...]
}
\end{lstlisting}